\documentclass[twoside]{article}

\usepackage{PRIMEarxiv}

\usepackage[utf8]{inputenc}     
\usepackage[T1]{fontenc}        
\usepackage{multicol}

\usepackage{microtype}          
\usepackage{setspace}           
\usepackage{textcomp}           
\usepackage{color,soul}         
\usepackage{lipsum}             

\usepackage{amsmath, amssymb, amsfonts, mathtools} 
\usepackage{eucal}              
\usepackage{mhchem}             
\usepackage{nicefrac}           
\usepackage{nccmath}

\usepackage{graphicx}           
\graphicspath{{media/}}         
\usepackage{caption, subcaption}
\usepackage{float}              
\usepackage{adjustbox}          
\usepackage{booktabs}           
\usepackage{multirow}           
\usepackage{longtable}          
\usepackage{rotating}           
\usepackage{pdflscape}          
\usepackage{placeins}           

\usepackage{algorithm}          
\usepackage{algpseudocode}      

\usepackage{enumitem}           

\usepackage{hyperref}           
\usepackage{url}                

\usepackage{fancyhdr}           

\usepackage{natbib}             

\usepackage{xpatch}             

\title{Physics-Informed Neural Networks with Hard Nonlinear Equality and Inequality Constraints
}

\author{
  Ashfaq Iftakher \thanks{Artie McFerrin Department of Chemical Engineering, Texas A\&M University, College Station, TX 77843-3122, USA} \\
  \texttt{iftakher@tamu.edu}
  \And
  Rahul Golder \footnotemark[1] \\
  \texttt{rahulgolder8420@tamu.edu}
  \AND
  Bimol Nath Roy \footnotemark[1]\\
  \texttt{bimolnathroy@tamu.edu}
  \And
  M. M. Faruque Hasan \footnotemark[1]   \thanks{Texas A\&M Energy Institute, Texas A\&M University, College Station, TX 77843, USA}   \thanks{Corresponding author: \texttt{hasan@tamu.edu}}\\
  \texttt{hasan@tamu.edu} \\
}

\begin{document}
\maketitle

\begin{abstract}

Traditional physics-informed neural networks (PINNs) do not guarantee strict constraint satisfaction. This is problematic in engineering systems where minor violations of governing laws can degrade the reliability and consistency of model predictions. In this work, we introduce KKT-Hardnet\begingroup\renewcommand\thefootnote{+}\footnotemark\footnotetext{Code available at \url{https://github.com/SOULS-TAMU/kkt-hardnet}}\endgroup{}, a neural network architecture that enforces linear and nonlinear equality and inequality constraints up to machine precision. It leverages a differentiable projection onto the feasible region by solving Karush-Kuhn-Tucker (KKT) conditions of a distance minimization problem. Furthermore, we reformulate the nonlinear KKT conditions via a log-exponential transformation to construct a sparse system with linear and exponential terms. We apply KKT-Hardnet to nonconvex pooling problem and a real-world chemical process simulation. Compared to multilayer perceptrons and PINNs, KKT-Hardnet achieves strict constraint satisfaction. It also circumvents the need to balance data and physics residuals in PINN training. This enables the integration of domain knowledge into machine learning towards reliable hybrid modeling of complex systems.

\end{abstract}


\section{Introduction}
Neural networks have gained widespread popularity as surrogate models that enable fast approximate predictions of complex physical systems. Among them, physics-informed neural networks (PINNs) have emerged as a powerful technique to embed first principles-based physical knowledge directly into the learning process. Unlike purely data-driven models, PINNs integrate known physical constraints into the training objective, typically by augmenting the loss function with penalty terms that account for constraint violations \cite{karniadakis2021physics}. These constraints may include algebraic equations \cite{iftakher2025integrating}, ordinary and partial differential equations \cite{eivazi2022physics, chen2021theory, raissi2020hidden}, or combinations thereof. Although such high-fidelity models mimic the underlying physics, they are often computationally prohibitive, particularly in practical scenarios, such as real-time control \cite{sanchez2018real} or large-scale optimization \cite{kumar2021industrial}. This has motivated the development of models that approximate the input–output behavior of these systems while reducing computational burden \cite{BHOSEKAR2018, iftakher2022data, iftakher2022guaranteed, Mcbride2019}. Surrogate models are increasingly used in chemical process systems, ranging from thermodynamic property predictions to unit operations and full process flowsheet simulation and optimization \cite{misener2023formulating}.

Among various surrogate modeling strategies, symbolic regression tools like ALAMO \cite{wilson2017alamo, cozad2014learning} optimize a linear combination of basis functions using mathematical programming, yielding interpretable functional forms. Artificial Neural networks (ANN) approximate continuous nonlinear functions \cite{hornik1989multilayer}, and have shown exceptional performance across a variety of engineering domains and tasks, including computer vision and language \cite{krizhevsky2017imagenet,goodfellow2020generative,lecun2015deep,vaswani2017attention}, model predictive control \cite{ren2022tutorial}, transport phenomena \cite{cai2021physicsa} and structure–property relationships for complex molecules \cite{iftakher2025multi}. However, the black-box nature of ANNs hinders their utility in domains that require physical consistency and interpretability \cite{kashinath2021physics}. The unconstrained ANN training often results in predictions that violate known conservation laws, making them unreliable for decision-making in safety-critical applications. To mitigate this, PINNs incorporate physical laws, typically as soft constraints, into the loss function. While soft-constrained PINNs are flexible and generalizable, they suffer from several drawbacks. First, this multi-objective optimization trades prediction accuracy with physical fidelity, but does not guarantee strict satisfaction of the constraints \cite{raissi2019physics, raissi2020hidden, cai2021physics}. Constraint violations (although penalized) can persist, especially when data is scarce or highly nonlinear \cite{chen2024physics, wang2022and}. This is particularly problematic in process systems engineering applications, where surrogate models are cascaded across interconnected unit operations. Even small constraint violations at the component level can accumulate and propagate, undermining the validity of the full system model \cite{ma2022data}. Second, training PINNs is challenging due to the nonconvexity of the objective function landscape induced by soft constraints and the requirement of careful tuning of the penalty parameters \cite{krishnapriyan2021characterizing, wang20222, rathore2024challenges}. 

To that end, hard-constrained ANNs that embed strict equality and inequality constraints into the model architecture have recently gained increasing attention. Enforcing hard physical constraints can help regularize models in low-data regimes and prevent overfitting \cite{min2024hard,marquez2017imposing}. Hard constrained machine learning approaches typically fall into two broad categories: \textit{projection-based} methods and \textit{predict-and-complete} methods. Projection methods correct errors due to unconstrained predictions by projecting them onto the feasible region defined by the constraints. This is achieved by solving a distance minimization problem\cite{chen2024physics, min2024hard}. In contrast, predict-and-complete architectures generate a subset of the outputs and analytically/numerically complete the rest using constraint equations \cite{beucler2021enforcing, donti2021dc3}. Both methods have demonstrated effectiveness in ensuring feasibility during both training and inference. Several notable frameworks have introduced differentiable optimization layers within ANNs to enforce constraints through the Karush-Kuhn-Tucker (KKT) conditions \cite{nocedal1999numerical}. OptNet \cite{amos2017optnet} pioneered this idea for quadratic programs, which was later extended to general convex programs with implicit differentiation \cite{agrawal2019differentiable}. These techniques have enabled ANNs to solve constrained optimization problems within the forward pass, allowing gradients to propagate through the solution map. Recent works such as HardNet-Cvx \cite{min2024hard} and DC3 \cite{donti2021dc3} have applied several of these ideas to a range of convex and nonlinear problems, leveraging fixed-point solvers and the implicit function theorem to compute gradients. For the special case of linear constraints, closed-form projection layers can be derived analytically, thereby avoiding any iterative schemes to ensure feasibility \cite{chen2021theory, chen2024physics}. Recently, Lastrucci and Schweidtmann \cite{lastrucci2025enforce} introduced ENFORCE that employs a scalable adaptive differentiable projection module to enforce a set of $\mathcal{C}^1$ nonlinear equality constraints. While most problems of practical importance are nonlinear, limited work exists to ensure hard nonlinear inequality and equality constraints in ANN outputs, especially those involving exponential, polynomial, multilinear, or rational terms in both the inputs and outputs. 


In this work, we introduce KKT-Hardnet, a neural network architecture that rigorously enforces hard nonlinear equality and inequality constraints in both input and output variables. KKT-Hardnet ensures constraint satisfaction by solving a square system that arises from the KKT condition of a distance minimization problem (i.e., projection problem). A Newton-type iterative scheme is embedded as a differentiable projection layer. The projection step ensures that the final output satisfies the original nonlinear constraints, with the unconstrained multilayer perceptron (MLP) prediction serving as the initial guess. To handle general nonlinearities without resorting to arbitrary basis expansions, we also demonstrate a symbolic transformation strategy, where nonlinear algebraic expressions are reformulated using a sequence of auxiliary variables, logarithmic transformations, and exponential substitutions. This results in a generalized system of linear and exponential terms. In our formulation, the slack and dual variables for inequality constraints are automatically guaranteed to be nonnegative. We demonstrate the proposed framework on illustrative examples and a pooling problem with nonlinear equality and inequality constraints in both inputs and outputs, as well as using a real-world chemical process simulation problem. Our results show that the proposed approach ensures constraint satisfaction within machine precision or specified tolerance. Embedding hard constraints improves both prediction fidelity and generalization performance when compared to unconstrained MLPs and PINNs with soft constraint regularization. 

\vspace{0.5em}
\noindent
To summarize, the main contributions of this work are as follows:
\begin{itemize}
    \item We develop a physics-informed neural network architecture, KKT-Hardnet, for hard constrained machine learning. KKT-Hardnet embeds a projection layer onto a neural net backbone to solve the KKT system of a projection problem via a Newton-type iterative scheme, thereby ensuring hard constraint satisfaction.
    \item We modify the inequalities to equalities using slack variables and model the complementarity conditions using Fischer-Burmeister reformulation \cite{jiang1997smoothed} to ensure nonnegativity of the slack variables and Lagrange multipliers for the inequality constraints.
    \item We transform general nonlinear equality and inequality constraints into a structure composed of only linear and exponential relations, using log-exponential transformation of nonlinear terms. The transformation allows to isolate all the nonlinearities in a specific linear and exponential form.
    \item We show that KKT-Hardnet improves surrogate model fidelity compared to vanilla MLPs and soft-constrained PINNs, for example, in applications involving nonlinear thermodynamic equations and process simulations.
\end{itemize}

The remainder of this paper is organized as follows. Section \ref{sec:methodology} presents the architecture of KKT-Hardnet, along with the mathematical formulation of the projection mechanism, the KKT system, and the log-exponential transformation. Section \ref{sec:numerical_experiments} presents numerical results for a set of illustrative examples, as well as a highly nonlinear chemical process simulation problem. Finally, Section \ref{sec:conclusions} provides concluding remarks and directions for future work.

\section{Methodology}
\label{sec:methodology}
In many physical systems, the relationship between input variables \(\boldsymbol{x}\) and output variables \(\boldsymbol{y}\) is governed by \textit{known} mass and energy conservation laws, inter-variable dependencies, and physical limitations on the operation. These can often be expressed as linear/nonlinear equality or inequality constraints. To rigorously embed such domain knowledge into learning, we present KKT-Hardnet, a neural network architecture that guarantees satisfaction of these constraints during both training and inference. 

\begin{figure}[ht!]
    \centering
    \includegraphics[scale=0.55]{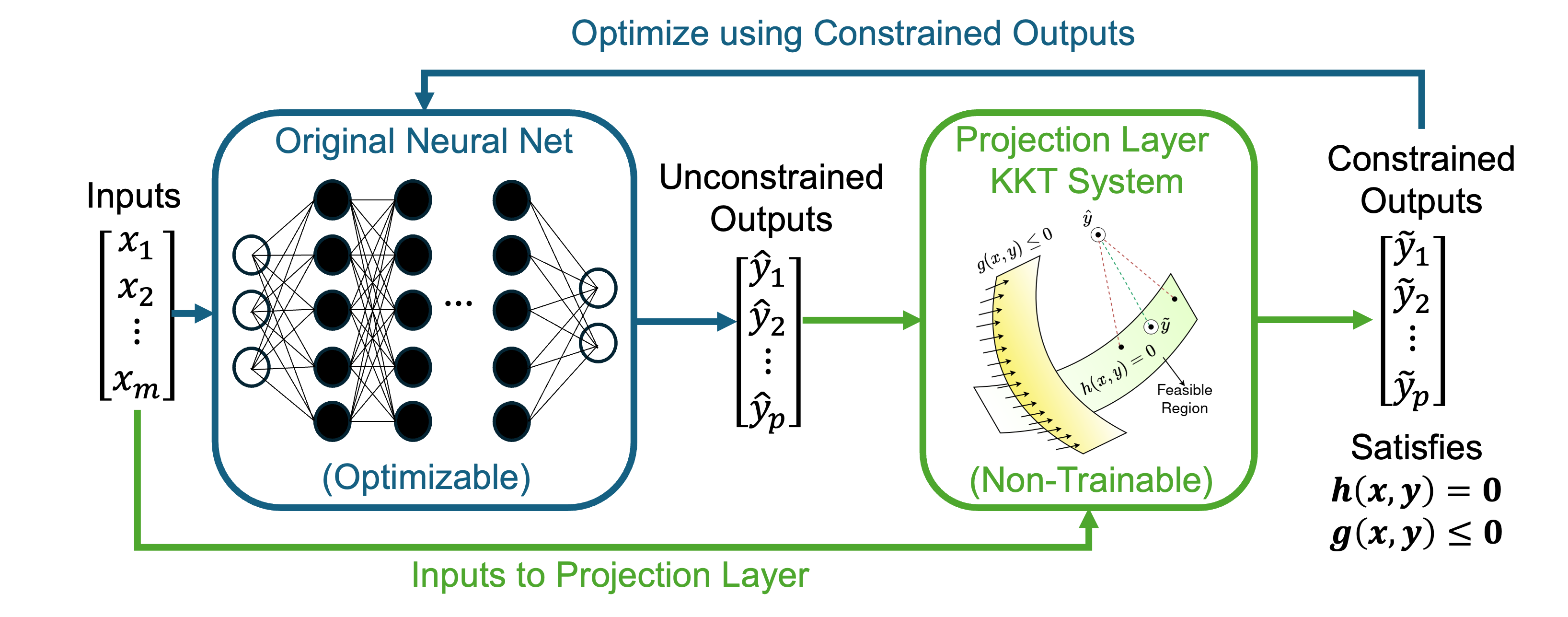}
    \caption{Neural network architecture of KKT-Hardnet for hard constrained machine learning. Unconstrained outputs are calculated using a standard neural net, while the constrained outputs are calculated as the solution of the nonlinear system of equations corresponding to the KKT relation of a distance minimization problem. The projection layer enforces raw neural network output to lie on the constraint manifold.}
    \label{pic-overview}
\end{figure}

Consider a dataset $\{(\boldsymbol{x}_i, \bar{\boldsymbol{y}}_i)\}_{i=1}^N$ representing the input–output behavior of a physical system. The objective is to train a neural network that approximates the underlying functional mapping between the inputs $\boldsymbol{x}$ and the outputs $\bar{\boldsymbol{y}}$. We consider that some prior knowledge about the system is also available in the form of algebraic constraints that relate the inputs $\boldsymbol{x}$ and the outputs $\bar{\boldsymbol{y}}$. Formally, the goal is to learn $\boldsymbol{y} = \mathrm{NN}(\boldsymbol{\Theta}, \boldsymbol{x})$ such that the predicted output $\boldsymbol{y}$ always belongs to the feasible set $\mathcal{S}$, i.e., $\boldsymbol{y} \in \mathcal{S}$, where $\mathcal{S}$ is defined as $\mathcal{S} = \{ \boldsymbol{y} | \boldsymbol{h(x,y) = 0}, \boldsymbol{g(x,y) \le 0} \}$. Specifically, $\boldsymbol{h(x,y) = 0}$ represents a set of algebraic equality constraints $h_k(\boldsymbol{x}, \boldsymbol{y}) = 0$ for $k \in \mathcal{N}_{E}$, and $\boldsymbol{g(x,y) \le 0}$ represents another set of algebraic inequality constraints $g_k(\boldsymbol{x}, \boldsymbol{y}) \le 0$ for $k \in \mathcal{N}_{I}$. The functions $\boldsymbol{h}$ and $\boldsymbol{g}$ encode domain-specific knowledge, and may be nonlinear in both $\boldsymbol{x}$ and $\boldsymbol{y}$. Here, all the aggregated tunable parameters (weights and biases) of an ANN are defined as $\boldsymbol{\Theta}$. We make the following assumptions:

\begin{enumerate}
    \item The number of equality constraints $\mathcal{N}_E$ is less than or equal to the number of outputs $p$ of the neural network, i.e., \( \mathcal{N}_E \leq p \), to ensure availability of sufficient degrees of freedom for learning from data while satisfying the constraints.
    \item The constraint functions $\boldsymbol{h}$ are feasible and linearly independent.
    \item The inputs are $\boldsymbol{x} \in \mathbb{R}^m$ and the outputs are ${\boldsymbol{y}} \in \mathbb{R}^p$.
\end{enumerate}

\begin{figure}[ht!]
    \centering
    \includegraphics[width=0.6\linewidth]{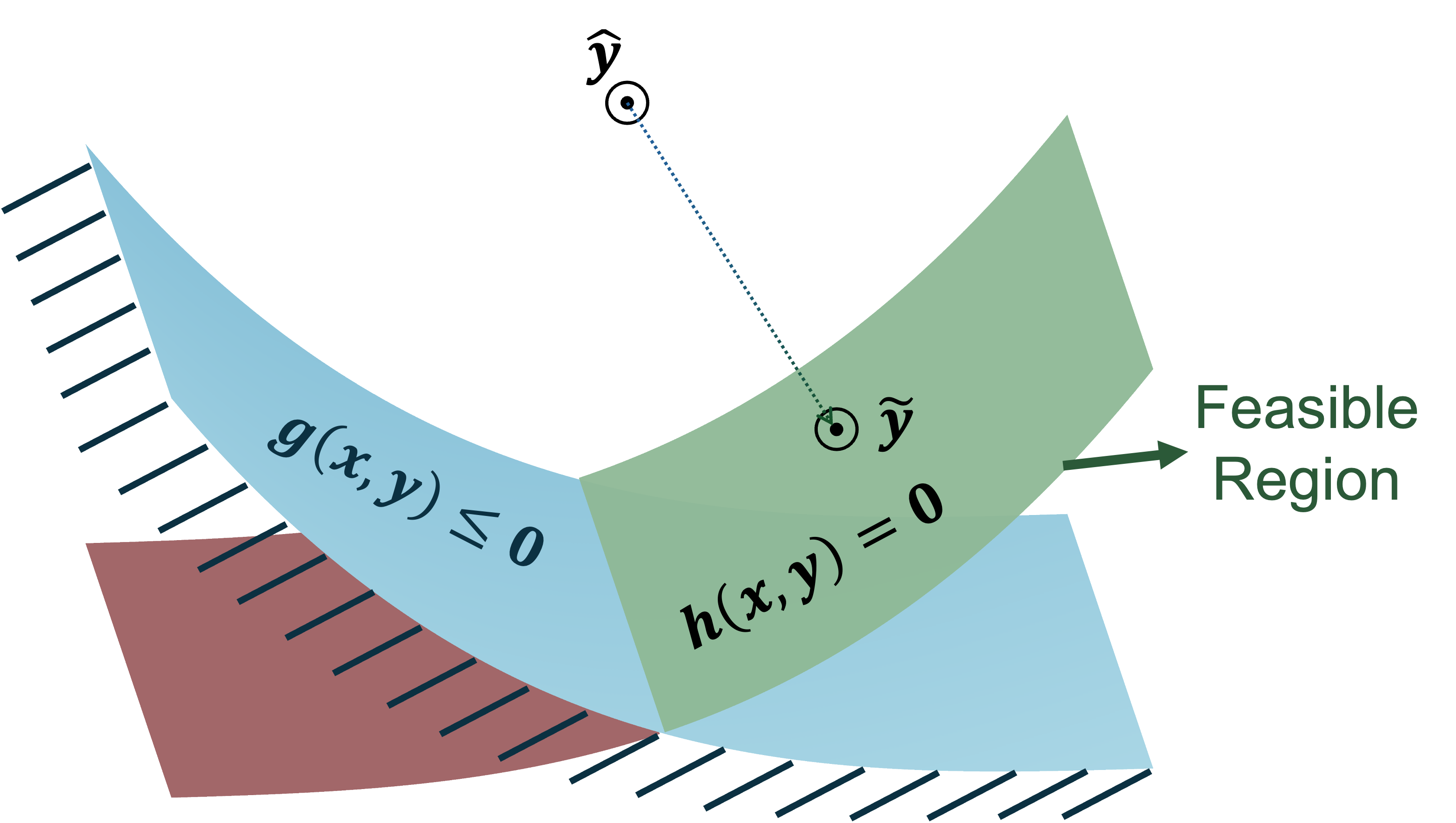}
    \caption{Projection of neural net outputs onto the constraint manifold.}
    \label{fig:projection}
\end{figure}

With these, our approach (see Figure \ref{pic-overview}) involves augmenting a standard neural network with a \textit{projection layer} acting as a corrector. Given an input \({\boldsymbol{x}}\), the network first computes an unconstrained output \(\hat{\boldsymbol{y}} = \mathrm{NN}({\boldsymbol{\Theta, x}})\), which may violate the physical constraints. The projection layer then adjusts this output to produce a corrected prediction \(\tilde{\boldsymbol{y}}\) that lies exactly on the constraint manifold (an illustration is shown in Figure \ref{fig:projection}). This strict enforcement of constraints is achieved by solving a square system of nonlinear equations derived from the KKT conditions associated with the following constrained minimization problem:

\begin{ceqn}
    \begin{equation}
    \label{eq-proj-general}
    \begin{aligned}
    \tilde{\boldsymbol{y}}^* =\; &\arg\min_{\boldsymbol{y}} \; \tfrac{1}{2} \left\| \boldsymbol{y} - \hat{\boldsymbol{y}} \right\|^2 \\
    &\quad \text{s.t.} \;\; h_k(\boldsymbol{x}, \boldsymbol{y}) = 0, \quad \forall k \in \mathcal{N}_E, \\
    &\quad\quad\;\; ~~ g_k(\boldsymbol{x}, \boldsymbol{y}) \le 0, \quad \forall k \in \mathcal{N}_I.
    \end{aligned}
    \end{equation}
\end{ceqn}

Essentially, this optimization problem searches for a \textit{minimum distance} solution from $\hat{\boldsymbol{y}}$ that lies on the feasible region defined by the equality and inequality constraints. The optimal solution $\tilde{\boldsymbol{y}}^*$ corresponds to a minimum distance projection of \(\hat{\boldsymbol{y}}\) onto the feasible region. Due to the presence of non-convexity, finding $\tilde{\boldsymbol{y}}^*$ requires a global optimization procedure, which may be computationally prohibitive. Therefore, we consider solving the system of nonlinear equations derived from the KKT conditions of \ref{eq-proj-general} that yield a feasible projection $\tilde{\boldsymbol{y}}$. For the special case of convexity, the KKT system is both necessary and sufficient to obtain the minimum distance solution $\tilde{\boldsymbol{y}}^*$. As we describe later, this feasible projection can be directly embedded in the forward pass of a neural network architecture through a Newton-type iterative scheme, ensuring that only physically consistent predictions are used for loss computation and backpropagation. As a result, the loss function for KKT-Hardnet is:

\begin{ceqn}
\begin{equation}
\mathcal{L}_{\mathrm{KKT-Hardnet}} = \frac{1}{2N} \sum_{i=1}^N \| \tilde{\boldsymbol{y}}_i - \bar{\boldsymbol{y}}_i \|^2,
\end{equation}
\end{ceqn}

\noindent in contrast to the conventional loss function used in standard neural networks:

\begin{ceqn}
\begin{equation}
\mathcal{L}_{\mathrm{NN}} = \frac{1}{2N} \sum_{i=1}^N \| \hat{\boldsymbol{y}}_i - \bar{\boldsymbol{y}}_i \|^2.
\end{equation}
\end{ceqn}

Notably, the use of the projected outputs \(\tilde{\boldsymbol{y}}\) alters the learning directions during training, which is informed by the feasible manifold. This ultimately guides the model toward parameters that fit the data while respecting the known constraints. This approach is similar to that of Chen et el. \cite{chen2024physics} applied to linear equalities. However, we extend it to accommodate both nonlinear equality and inequality constraints. KKT-Hardnet is architecture-agnostic and can be integrated with any neural network backbone, including convolutional neural networks (CNNs) or recurrent neural networks (RNNs).

To enforce the inequality constraints via projection, we introduce non-negative slack variable $s_k$ for $k  \in \mathcal{N}_{I}$ in the optimization problem described by Eq. \ref{eq-proj-general}:

\begin{ceqn}
\begin{equation}
\label{eq:ineq_proj_general}
\begin{aligned}
\tilde{\boldsymbol{y}}^* =\; \arg \min_{\boldsymbol{y}} \; &\tfrac{1}{2} \left\| \boldsymbol{y} - \hat{\boldsymbol{y}} \right\|^2 \\
\text{s.t.} \;\;
&h_k(\boldsymbol{x}, \boldsymbol{y}) = 0, \quad \forall k \in \mathcal{N}_E, \\
& g_k(\boldsymbol{x}, \boldsymbol{y}) + s_k = 0, \quad \forall k \in \mathcal{N}_I,\\
& s_k \ge 0, \quad \forall k \in \mathcal{N}_I.
\end{aligned}
\end{equation}
\end{ceqn}

\noindent We define the Lagrange multipliers \(\mu_k^{E} \in \mathbb{R}\), \(\mu_k^{I}\ge0\) for equality and inequality constraints respectively, and write the KKT conditions of the problem in Eq. \ref{eq:ineq_proj_general} as follows:

\begin{align}
&\text{(Stationarity)}
  &&\boldsymbol{y}-\hat{\boldsymbol{y}}
  + \sum_{k \in \mathcal{N}_E}\mu_k^{E}\,\nabla_{\boldsymbol{y}}h_k(\boldsymbol{x},\boldsymbol{y}) + \sum_{k \in \mathcal{N}_I}\mu_k^{I}\,\nabla_{\boldsymbol{y}}g_k(\boldsymbol{x},\boldsymbol{y}) = \boldsymbol{0},
  \label{eq:stat_general}\\[4pt]
&\text{(Primal feasibility)}\quad
  &&h_k(\boldsymbol{x},\boldsymbol{y}) = 0, \quad k \in \mathcal{N}_E
  \label{eq:prim_general_h}\\[4pt]
&\text{(Primal feasibility)}\quad
  &&g_k(\boldsymbol{x},\boldsymbol{y}) + s_k = 0, \quad k \in \mathcal{N}_I
  \label{eq:prim_general_g}\\[4pt]
&\text{(Dual feasibility)}\quad
  &&\mu_k^I \;\ge\;0,\quad s_k\;\ge\;0, \quad k \in \mathcal{N}_I
  \label{eq:dual_general}\\[4pt]
&\text{(Complementarity)}\quad
  &&\mu_k^I \cdot s_k 
  \;=\; 0, \quad k \in \mathcal{N}_I.
  \label{eq:comp_general}
\end{align}

\noindent Furthermore, instead of using original dual feasibility $\mu_k^I \ge0, s_k\ge0$ and complementarity condition $\mu_k^I\,s_k = 0$, we replace both conditions by the equivalent, Fischer–Burmeister reformulation \cite{jiang1997smoothed} as follows.

\begin{ceqn}
\begin{equation}
    \phi_k (\mu_k^I , s_k ) :\quad \mu_k^I + s_k - \sqrt{(\mu_k^{I})^2 + s_k^2}
  \;=\; 0, \quad k \in \mathcal{N}_I.
  \label{eq:fischer-burmeister}
\end{equation}
\end{ceqn}

Note that the Fischer-Burmeister reformulation automatically maintains nonnegativity of the dual and slack variables \(\mu_k^I \ge 0,\;s_k \ge 0\) during all iterations. With this reformulation, we are able to generalize the first-order KKT conditions in a compact form as follows:

\begin{ceqn}
\begin{equation}
\label{eq:KKT_formulation1}
\begin{aligned}
\quad (F_1)\;&\;
\boldsymbol{y}-\boldsymbol{\hat{y}}+ \nabla_{\boldsymbol{y}} \boldsymbol{\tilde{h}(\boldsymbol{x,y, \lambda})}^{\mathsf T}  \boldsymbol{\lambda}=0\\[2pt]
(F_2)\;&\;
 \boldsymbol{\tilde{h}(\boldsymbol{x,y, \lambda})}= \boldsymbol{0}\\[2pt]
\end{aligned}
\end{equation}
\end{ceqn}

\noindent where, $\boldsymbol{\lambda}$ is a vector created by concatenation:  $\boldsymbol{\lambda} = [\boldsymbol{\mu}^E \quad \boldsymbol{\mu}^I \quad \boldsymbol{s}]$, and $\boldsymbol{\tilde{h}}$ is a vector of functions created by concatenation: $\boldsymbol{\tilde{h}} = [\boldsymbol{h} \quad \boldsymbol{g + s} \quad \boldsymbol{\phi}].$ Therefore, $\tilde{h}_k = 0$ for $k \in \mathcal{N}_T$, where  $\mathcal{N}_T$ is the union of constraints given in Eqs. \ref{eq:prim_general_h}, \ref{eq:prim_general_g} and \ref{eq:fischer-burmeister}. This completes the first transformation of the KKT conditions as a system of (nonlinear) equalities. Note that Eq. \ref{eq:KKT_formulation1} is a square system of equations. Therefore, iterative methods can be directly applied to solve this system of nonlinear equations.

From here, a natural approach is to enforce these KKT conditions by employing an iterative Gauss-Newton procedure. The overall training and inference of KKT-Hardnet are described in Algorithm \ref{alg:KKT-Hardnet}.

\begin{algorithm}[H]
\caption{KKT-Hardnet: Nonlinear Equality and Inequality Constraint Satisfaction via Differentiable Projection}
\label{alg:KKT-Hardnet}
\begin{algorithmic}[1]
\Require Dataset $\boldsymbol{\mathcal D}=\{(\boldsymbol x, \boldsymbol{\bar{y}} \}$; neural net backbone $NN_{\boldsymbol{\Theta}}:\mathbb R^{m}\!\to\!\mathbb R^{p}$; projection routine $\rho(\cdot)$; Equality/inequality constraints as $\boldsymbol{h(x,y)}=0$, $\boldsymbol{g(x,y)}\le 0$; the KKT system corresponding to the projection problem (Eq. \ref{eq:KKT_formulation1}).


\vspace{0.6em}

\Procedure{Train}{$\boldsymbol{\mathcal D}$}
  \State initialize neural network $NN_{\boldsymbol{\Theta}}$
  \While{not converged}
    \For{minibatch $\boldsymbol{\mathcal B}\subset\boldsymbol{\mathcal D}$}
      \For{$(\boldsymbol x,\boldsymbol y)\in \boldsymbol{\mathcal B}$}
        \State \textbf{predict (unconstrained):} $\hat{\boldsymbol y}\gets NN_{\boldsymbol{\Theta}}(\boldsymbol x)$
        \State \textbf{project to feasible:} $\tilde{\boldsymbol y}\gets\rho(\hat{\boldsymbol y})$
        \State \textbf{compute loss:} $\mathcal{L}\big(\tilde{\boldsymbol y},\boldsymbol y\big)$ 
      \EndFor
      \State \textbf{update} $\boldsymbol{\Theta}$ using $\boldsymbol{\nabla_\Theta} \mathcal{L}\big(\tilde{\boldsymbol y},\boldsymbol y\big)$ by backpropagation
    \EndFor
  \EndWhile
\EndProcedure
\vspace{0.4em}

\Procedure{Test}{$\boldsymbol x, NN_{\boldsymbol{\Theta}}$}
  \State $\hat{\boldsymbol y}\gets NN_{\boldsymbol{\Theta}}(\boldsymbol x)$
  \State $\tilde{\boldsymbol y}\gets\rho(\hat{\boldsymbol y})$
  \State \Return $\tilde{\boldsymbol y}$
\EndProcedure
\vspace{0.6em}

\State \textbf{Note.}
\Statex \quad $\rho(\cdot)$ may be (i) a closed-form affine projector when constraints are linear in outputs, or (ii) Gauss--Newton steps on the KKT/log--exp system. 
\end{algorithmic}
\end{algorithm}

\subsection{Implementation of KKT-Hardnet}
\label{sec:kkt-implement}

Define:

\[
  \boldsymbol{F}(\boldsymbol{{x},\hat{y},y,\lambda}) \;=\; 
  \begin{pmatrix}
    F_{1}(\boldsymbol{{x},\hat{y},y,\lambda})\\[6pt]
    F_{2}(\boldsymbol{{x},y,\lambda})\\[6pt]
  \end{pmatrix}
  = 0
\]

where, $F_1$ and $F_2$ are described in Eq. \ref{eq:KKT_formulation1}. The Jacobian with respect to $(\boldsymbol{y, \lambda})$ is

\[
\boldsymbol{J(y,\lambda)}
:=
\frac{\partial \boldsymbol{F}}{\partial(\boldsymbol{y,\lambda})}.
\]

At iteration $k$, we solve the Newton step when the Jacobian is invertible as follows:

\[
\boldsymbol{J}^{(k)}
\begin{bmatrix}
  \Delta \boldsymbol{y}\\[4pt]
  \Delta \boldsymbol{\lambda}
\end{bmatrix}
=
-\,\boldsymbol{F}^{(k)},
\]

A regularized Gauss-Newton step is also available for the case when the Jacobian is noninvertible:

\[
\bigl[{\boldsymbol{J}^{(k)\top}} \boldsymbol{J}^{(k)} + \gamma\boldsymbol{I}\bigr]
\begin{bmatrix}
  \Delta \boldsymbol{y}\\[4pt]
  \Delta \boldsymbol{\lambda}
\end{bmatrix}
=
-\,\boldsymbol{J}^{(k)\top}\boldsymbol{F}^{(k)},
\]

where, $\boldsymbol{J}^{(k)} = \boldsymbol{J}(\boldsymbol{y}^{(k)},\boldsymbol{\lambda}^{(k)})$ and $\boldsymbol{F}^{(k)} = \boldsymbol{F}(\boldsymbol{{x}},\boldsymbol{\hat{y}},\boldsymbol{y}^{(k)},\boldsymbol{\lambda}^{(k)})$. Then update

\[
\begin{aligned}
  \boldsymbol{y}^{(k+1)}        &= \boldsymbol{y}^{(k)}        + \alpha\,\Delta \boldsymbol{y},\\[4pt]
  \boldsymbol{z}^{(k+1)}        &= \boldsymbol{z}^{(k)}        + \alpha\,\Delta \boldsymbol{z},\\[4pt]
  \boldsymbol{\lambda}^{(k+1)} &= \boldsymbol{\lambda}^{(k)} + \alpha\,\Delta \boldsymbol{\lambda},
\end{aligned}
\]

where \(\alpha\in(0,1]\) is a step‐length parameter (e.g.\ from Armijo line‐search) and \(\gamma>0\) is a small Tikhonov‐regularization parameter to ensure invertibility.

During training, we optionally employ an adaptive “warm start” for the Gauss–Newton projection layer. We first train the backbone network alone and monitor the data loss $\mathcal{L}_{\mathrm{NN}}$. When this loss falls below a user–specified threshold $\eta$ (or after a fixed warm-up budget), we enable the projection layer and continue end-to-end training. Adaptive activation of the projection improves the initial guess $\hat{\boldsymbol y}$ supplied to the Gauss–Newton solve, which may reduce the number of projection iterations to converge to the user-specified tolerance.

\paragraph{Differentiation through the projection layer.} Starting with $\boldsymbol{y}^{(0)}=\hat{\boldsymbol{y}}$, the projection layer runs a fixed number (\(K\)) of Newton/Gauss--Newton steps to enforce the algebraic constraints. During backpropagation, these \(K\) iterations are unrolled in the computation graph. Automatic differentiation propagates 
\(\partial\mathcal L/\partial \tilde{\boldsymbol y}\) backwards through each update 
\(\boldsymbol y^{(k+1)}=\phi(\boldsymbol y^{(k)})\) to obtain 
\(\mathrm d\mathcal L/\mathrm d\boldsymbol\Theta\) using chain rules. Although unrolling is simple, it is memory-heavy. This is because all intermediate Newton/Gauss-Newton iterates 
\(\{\boldsymbol y^{(k)}\}_{k=0}^{K}\) are stored for the backward pass. For a more memory-efficient implementation, one can reduce \(K\), or switch to the implicit/VJP formulation 
outlined in Appendix~B, which needs only a single adjoint solve and does not store all intermediate iterates.

\subsection{Log-Exponential Transformation}
\label{subsec:log_transformation}

A symbolic transformation strategy via logarithmic and exponential substitutions may allow general nonlinear constraints to be rewritten in a structured form where all nonlinearities appear as exponentials and/or linear combinations of variables. Specifically, any constraint of the form $h(\boldsymbol{x}, \boldsymbol{y}) = 0$, that may include nonlinear terms such \( y^n \), \( xy \), or \( y/x\) and so on, can be reformulated into an equivalent system
where \(\boldsymbol{z}\) denotes a set of new auxiliary variables that allows each nonlinear term to be replaced by an exponential. 
This yields a constraint system of the general form which is equivalent to the system given by Eq. \ref{eq:KKT_formulation1}:

\begin{ceqn}
\begin{subequations}
\label{eq:Primal-feasibility_eq}
\begin{align}
\boldsymbol{Ax} + \boldsymbol{By} + \boldsymbol{A_x} \exp({\boldsymbol{x}}) + \boldsymbol{C_z}{\boldsymbol{z}} + \boldsymbol{C_{\lambda}}{\boldsymbol{\lambda}} &= \boldsymbol{b}, \label{eq:log_e1} \\
\quad \boldsymbol{D_{y}y} + \boldsymbol{D_{z}z} +  \boldsymbol{D_{\lambda}\lambda} &= \boldsymbol{d}, \label{eq:log_e2} \\
\quad \boldsymbol{E_{y}y} + \boldsymbol{E_{z}z} + \boldsymbol{E_{\lambda}\lambda} &= \boldsymbol{G}\exp({\boldsymbol{H_{y}y + H_{z}z}}). \label{eq:log_e3}
\end{align}
\end{subequations}
\end{ceqn}

\noindent In general, we can denote this new system of equations in Eq. \ref{eq:Primal-feasibility_eq} as $\boldsymbol{F}(\boldsymbol{y}, \boldsymbol{z}, \boldsymbol{\lambda}) = \mathbf{0}$. The solution of this system of equations gives $\boldsymbol{\tilde{y}}$. Note that, in this transformed general form, all terms are linear except for $\exp({\boldsymbol{H_{y}y + H_{z}z}})$ in Eq. \ref{eq:log_e3}.

To illustrate, consider the nonlinear algebraic constraint:

\begin{ceqn}
\begin{equation}
\label{eq:example_constraint}
h(\boldsymbol{x},\boldsymbol{y}) := y_1 - y_2^{3} - 12x^{2} + 6x - 6 = 0.
\end{equation}
\end{ceqn}

\noindent Let $z_1 := e^{z_5}, z_2 := e^{z_4}, z_3 := y_2^3, \quad z_4 := \log z_3, \quad z_5 := \log y_2$. This implies $z_4 = 3z_5, \quad y_2 = e^{z_5}, \quad z_3 = e^{z_4}$. For inputs, we define: $x_1 := x, \quad x_2 := x^2, \quad x_3 := \log x_2, \quad x_4 := \log x_1$, which implies: $x_3 = 2x_4, \quad x_2 = e^{x_3}, \quad x_1 = e^{x_4}$.

Using these definitions and auxiliary variables, the original constraint \eqref{eq:example_constraint} can be rewritten as the following system:

\begin{ceqn}
\begin{subequations}
\label{eq:full_linexp}
\begin{align}
y_1 - z_3 - 12x_2 + 6x_1 &=6, \label{eq:e1} \\
x_3 - 2x_4 &= 0,               \label{eq:e5} \\
x_2 - e^{x_3} &= 0,            \label{eq:e6} \\
x_1 - e^{x_4} &= 0.            \label{eq:e7} \\
z_4 - 3z_5 &= 0,               \label{eq:e2} \\
y_2 - z_1 &= 0,            \label{eq:e3} \\
z_3 - z_2 &= 0,            \label{eq:e4} \\
z_1 - e^{z_5} &=0,              \label{eq:e8} \\
z_2 - e^{z_4} &=0,              \label{eq:e9}
\end{align}
\end{subequations}
\end{ceqn}

where, $\boldsymbol{x} = (x_1, x_2, x_3, x_4)^\top$, $\boldsymbol{y} = (y_1,y_2)^\top$, and $\boldsymbol{z} = (z_1,\cdots,z_5)^\top$, with:


\begin{multicols}{2}
\noindent
$\boldsymbol{A}=
\begin{bmatrix}
6 & -12 & 0 & 0 \\
0 & 0 & 1 & -2 \\
0 & 1 & 0 & 0 \\
1 & 0 & 0 & 0
\end{bmatrix}$

\vspace{1em}

$\boldsymbol{B}=
\begin{bmatrix}
1 & 0 \\
0 & 0 \\
0 & 0 \\
0 & 0
\end{bmatrix}$

\vspace{1em}

$\boldsymbol{A}_x=
\begin{bmatrix}
0 & 0 & 0 & 0 \\
0 & 0 & 0 & 0 \\
0 & 0 & -1 & 0 \\
0 & 0 & 0 & -1
\end{bmatrix}$

\vspace{1em}

$\boldsymbol{C}_z=
\begin{bmatrix}
0 & 0 & -1 & 0 & 0 \\
0 & 0 & 0 & 0 & 0 \\
0 & 0 & 0 & 0 & 0 \\
0 & 0 & 0 & 0 & 0
\end{bmatrix}$

\vspace{1em}

$\boldsymbol{b}=
\begin{bmatrix}
6\\ 0\\ 0\\ 0
\end{bmatrix}$

\vspace{1em}

$\boldsymbol{D}_y=
\begin{bmatrix}
0 & 0 \\
0 & 1 \\
0 & 0
\end{bmatrix}$

\vspace{1em}

$\boldsymbol{D}_z=
\begin{bmatrix}
0 & 0 & 0 & 1 & -3 \\
-1 & 0 & 0 & 0 & 0\\
0 & -1 & 1 & 0 & 0
\end{bmatrix}$

\vspace{1em}

$\boldsymbol{d}=
\begin{bmatrix}
0\\ 0\\ 0
\end{bmatrix}$

\vspace{1em}

$\boldsymbol{E}_z=
\begin{bmatrix}
1 & 0 & 0 & 0 & 0\\
0 & 1 & 0 & 0 & 0
\end{bmatrix}$

\vspace{1em}

$\boldsymbol{H}_z=
\begin{bmatrix}
0 & 0 & 0 & 0 & -1 \\
0 & 0 & 0 & -1 & 0
\end{bmatrix}$

\vspace{1em}

$\boldsymbol{G}=
\begin{bmatrix}
1 & 0 \\
0 & 1
\end{bmatrix}$

\vspace{1em}

$\boldsymbol{H}=
\begin{bmatrix}
0 & 0 & 0 & 0 & 1\\
0 & 0 & 0 & 1 & 0
\end{bmatrix}$
\end{multicols} 
All other vectors and matrices are zero.

This transformation has several advantages. First, it maps a broad class of algebraic nonlinearities (e.g., products, powers, ratios) into a sparse “linear $+$ exponential” structure that is well suited to Newton-type solvers. Jacobians can easily be computed, and sparsity can be exploited. Second, it enables hard enforcement of nonlinear constraints by solving a structured system with only linear and exponential terms, without the need for an arbitrary number of basis functions. Third, the log–exponential reparameterization is monotone and preserves nonnegativity for selected variables (e.g., slacks and inequality multipliers), which helps maintain feasibility during line search and step selection, often improving robustness in practice. When combined with the KKT reformulation, the log–exponential parameterization naturally enforces nonnegativity of inequality slacks and multipliers, which eliminates the need for active-set type strategies. Also, the “linear $+$ exponential” structure leads to a Jacobian where the only nonlinear blocks are diagonal matrices of exponentials.

\vspace{0.5 cm}

\textit{\underline{Remark 1}.} Log-exponential transformation is applicable for systems with nonnegative outputs. That is,  the inputs are $\boldsymbol{x} \in \mathbb{R}^m$ and the outputs are ${\boldsymbol{y}} \in \mathbb{R}_{\ge 0}^p$.

\textit{\underline{Remark 2}.} We interpret all exponentials component‐wise.  That is, for any vector $\boldsymbol{v}$,  $\exp(\boldsymbol{v}) =\;\bigl[e^{v_1},\,e^{v_2},\dots\bigr]^{\!\top}$. In particular,
$\exp(\boldsymbol{H_{y}y})
  \;=\;
  \begin{bmatrix}
    e^{(\boldsymbol{H_{y}\,y})_1}\\
    e^{(\boldsymbol{H_{y}\,y})_2}\\
    \vdots
  \end{bmatrix},
$ where each entry is the exponential of the corresponding row of $\boldsymbol{H_{y}\,y}$.  

\textit{\underline{Remark 3}.}
Lagrange multipliers associated with equality constraints are unrestricted in sign. Therefore, applying a log--exponential transformation to those multipliers forces $\mu^E\ge 0$ and can render the KKT system inconsistent or ill-conditioned in pathological cases. 
We consider the following practical options:

\begin{enumerate}[label=(\roman*), leftmargin=1.6em]
\item Do not log--exp the equality multipliers. Specifically, ensure $\mu^E\in\mathbb{R}$ by applying the log--exp transformation only to the \emph{primal--dual feasibility} conditions and inequality slacks and multipliers, but leave the stationarity equation (where $\mu^E$ appears) in its original signed form.

\item Rewrite $\boldsymbol{h(x,y)}=0$ as the pair $\boldsymbol{h(x,y)}\le 0$ and $-\boldsymbol{h(x,y)}\le 0$. Then introduce nonnegative slacks, and enforce complementarity with a Fischer--Burmeister reformulation. This yields nonnegative multipliers throughout the KKT system.

\item Represent each equality multiplier as a difference of two nonnegative parts, 
$\mu^E=\mu^+ - \mu^-$ with $\mu^+,\mu^-\ge 0$, and (optionally) apply log--exp to $\mu^+,\mu^-$. 
This preserves unrestrictedness in sign while keeping nonnegativity at the auxiliary variable definitions. However, it increases variables and may be less stable than (i).
\end{enumerate}

\noindent
We adopt option~(i) for numerical experiments involving log-exp transformation unless stated otherwise.



\paragraph{Numerical solution of the log-exponential transformed system of the general KKT conditions.}
As described above, the log-exponential transformed system corresponding to the  KKT conditions for general linear/nonlinear equality and inequality constraints can be represented as Eq. \ref{eq:Primal-feasibility_eq}. Now define:

\[
  \mathbf{F}(\boldsymbol{y,z,\lambda}) \;=\; 
  \begin{pmatrix}
    F_{1}(\boldsymbol{y,z,\lambda})\\[6pt]
    F_{2}(\boldsymbol{y,z,\lambda})\\[6pt]
    F_{3}(\boldsymbol{y,z,\lambda})
  \end{pmatrix}
  = 0
\]
where
\begin{align}
F_{1}(\boldsymbol{y,z,\lambda})
&:= \boldsymbol{A\,\hat{x}}
  \;+\boldsymbol{\;B\,y
  \;}+\boldsymbol{\;A_{x}\,}\exp(\boldsymbol{\hat{x}})
  \;+\boldsymbol{\;C_{z}\,z
  \;}+\boldsymbol{\;C_{\lambda}\,\lambda
  \;}-\boldsymbol{\;b}, \nonumber\\[4pt]
F_{2}(\boldsymbol{y,z,\lambda})
&:= \boldsymbol{D_{y}\,y
  \;}+\boldsymbol{\;D_{z}\,z
  \;}+\boldsymbol{\;D_{\lambda}\,\lambda
  \;}-\boldsymbol{\;d}, \nonumber\\[4pt]
F_{3}(\boldsymbol{y,z,\lambda})
&:= \boldsymbol{E_{y}\,y
  \;}+\boldsymbol{\;E_{z}\,z
  \;}+\boldsymbol{\;E_{\lambda}\,\lambda
  \;}-\boldsymbol{G}\exp({\boldsymbol{H_{y}y + H_{z}z}}).
\label{eq:F_general}
\end{align}

The Jacobian with respect to \((\boldsymbol{y,z,\lambda})\) is

\begin{equation}
\boldsymbol{J(y,z,\lambda)}
:=
\frac{\partial \mathbf{F}}{\partial(\boldsymbol{y,z,\lambda})}
=
\begin{pmatrix}
  \boldsymbol{B} & \boldsymbol{C}_{\boldsymbol{z}} & \boldsymbol{C}_{\boldsymbol{\lambda}} \\[6pt]
  \boldsymbol{D}_{\boldsymbol{y}} & \boldsymbol{D}_{\boldsymbol{z}} & \boldsymbol{D}_{\boldsymbol{\lambda}} \\[6pt]
  \boldsymbol{E}_{\boldsymbol{y}} - \boldsymbol{G} \boldsymbol{L} \boldsymbol{H}_{\boldsymbol{y}} &
  \boldsymbol{E_z} - \boldsymbol{G} \boldsymbol{L} \boldsymbol{H}_{\boldsymbol{z}} &
  \boldsymbol{E}_{\boldsymbol{\lambda}}
\end{pmatrix},
\label{eq:J_general}
\end{equation}

\noindent
where
\[
\boldsymbol{L} := \mathrm{diag}\!\left(e^{\boldsymbol{H_y y + H_z z}}\right).
\]

With this Jacobian, we then apply the Newton/Gauss-Newton procedure (see Section \ref{sec:kkt-implement} for details).

\vspace{1 cm}

\noindent \textbf{Special Case 1: Equality constraints with nonlinearity in \(\boldsymbol{x}\) but
               linearity in \(\boldsymbol{y}\)}
\label{sec:case2}

If the equality constraints are affine in \(\boldsymbol{y}\), then one does not require Newton-type iterative scheme and can derive the projection \textit{analytically} as follows. First, we solve the simplified quadratic program to minimize the Euclidean distance between $\boldsymbol{\hat{y}}_0$ and $\tilde{\boldsymbol{y}}$ that lies on the constraint manifold.  
\begin{ceqn}
\begin{equation}
\begin{aligned}
\label{GC}
\tilde{\boldsymbol{y}} =~&\arg \min_{\boldsymbol{y}} \hspace{0.1 cm} \frac{1}{2} \| \boldsymbol{y} - \boldsymbol{\hat{y}}_0 \|^2, \\ 
&~\mbox{s.t.} \hspace{0.3 cm} \boldsymbol{A} \boldsymbol{\hat{x}} + \boldsymbol{B} \boldsymbol{y} + \boldsymbol{A_{x}} \exp(\boldsymbol{\hat{x}}) = \boldsymbol{b},\\ 
	\end{aligned} 
\end{equation} 
\end{ceqn}

The Lagrangian can then be constructed as follows:

\begin{ceqn}
\begin{equation}
\mathcal{L}(\boldsymbol{y}, \boldsymbol{\lambda}) = \frac{1}{2} (\boldsymbol{y} - \boldsymbol{\hat{y}}_0)^T (\boldsymbol{y} - \boldsymbol{\hat{y}}_0) + \boldsymbol{\lambda}^{\top}(\boldsymbol{A} \boldsymbol{\hat{x}} + \boldsymbol{B} \boldsymbol{y} + \boldsymbol{A_{x}} \exp(\boldsymbol{\hat{x}}) - \boldsymbol{b}) 
\end{equation}
\end{ceqn}

Stationary condition:

\begin{ceqn}
\begin{equation}
\frac{\partial \mathcal{L}}{\partial \boldsymbol{y}} = (\boldsymbol{y} - \boldsymbol{\hat{y}}_0) + \boldsymbol{B}^T \boldsymbol{\lambda} = 0 
\Rightarrow \boldsymbol{y} = \boldsymbol{\hat{y}}_0 - \boldsymbol{B}^T \boldsymbol{\lambda}
\end{equation}
\end{ceqn}

Dual feasibility:

\begin{ceqn}
\begin{equation}
\begin{aligned}
\frac{\partial \mathcal{L}}{\partial \boldsymbol{\lambda}} &= \boldsymbol{A} \hat{\boldsymbol{x}} + \boldsymbol{B} \boldsymbol{y} + \boldsymbol{A_{x}} \exp(\hat{\boldsymbol{x}}) - \boldsymbol{b} = 0 \\
&\Rightarrow \boldsymbol{A} \hat{\boldsymbol{x}} + \boldsymbol{B} (\boldsymbol{\hat{y}}_0 - \boldsymbol{B}^T \boldsymbol{\lambda}) + \boldsymbol{A_{x}} \exp (\hat{\boldsymbol{x}}) - \boldsymbol{b} = 0 \\ 
&\Rightarrow \boldsymbol{A} \hat{\boldsymbol{x}} + \boldsymbol{B} \boldsymbol{\hat{y}}_0 + \boldsymbol{A_{x}} \exp (\hat{\boldsymbol{x}}) - \boldsymbol{b} = \boldsymbol{B} \boldsymbol{B}^T \boldsymbol{\lambda}
\end{aligned}
\end{equation}
\end{ceqn}

If $\boldsymbol{B}$ is full rank, then $\boldsymbol{B} \boldsymbol{B}^T$ is invertible.

\begin{ceqn}
\begin{equation}
\begin{aligned}
\boldsymbol{\lambda} = (\boldsymbol{B} \boldsymbol{B}^T)^{-1} \left( \boldsymbol{B} \boldsymbol{\hat{y}}_0 + \boldsymbol{A} \hat{\boldsymbol{x}} + \boldsymbol{A_{x}} \exp \hat{\boldsymbol{x}} - \boldsymbol{b} \right)
\end{aligned}
\end{equation}
\end{ceqn}

\begin{ceqn}
\begin{equation}
\label{Eq-Analytic-Nonlinear-x}
\begin{aligned}
\tilde{\boldsymbol{y}} &= \boldsymbol{\hat{y}}_0 - \boldsymbol{B}^T (\boldsymbol{B} \boldsymbol{B}^T)^{-1} \left( \boldsymbol{B} \boldsymbol{\hat{y}}_0 + \boldsymbol{A} \hat{\boldsymbol{x}} + \boldsymbol{A_{x}} \exp(\hat{\boldsymbol{x}}) - \boldsymbol{b} \right) \\
&= \left[ \boldsymbol{I} - \boldsymbol{B}^T (\boldsymbol{B} \boldsymbol{B}^T)^{-1} \boldsymbol{B} \right] \boldsymbol{\hat{y}}_0 \\
&\quad + \left[ \boldsymbol{B}^T (\boldsymbol{B} \boldsymbol{B}^T)^{-1} \boldsymbol{A} \right] \hat{\boldsymbol{x}} \\
&\quad + \left[ -\boldsymbol{B}^T (\boldsymbol{B} \boldsymbol{B}^T)^{-1} \boldsymbol{A_{x}} \right] \exp(\hat{\boldsymbol{x}}) \\
&\quad + \boldsymbol{B}^T (\boldsymbol{B} \boldsymbol{B}^T)^{-1} \boldsymbol{b} \\
&= \boldsymbol{A}^* \hat{\boldsymbol{x}} + \boldsymbol{B}^* \boldsymbol{\hat{y}}_0 + \boldsymbol{A_{x}}^* \exp(\hat{\boldsymbol{x}}) + \boldsymbol{b}^*
\end{aligned}
\end{equation}
\end{ceqn}

where:

\begin{ceqn}
\begin{subequations}\label{eq:projection-block}
\begin{align}
\boldsymbol{A}^*     &= \boldsymbol{B}^\top (\boldsymbol{B} \boldsymbol{B}^\top)^{-1} \boldsymbol{A}, \label{eq:proj-Astar} \\
\boldsymbol{B}^*     &= \boldsymbol{I} - \boldsymbol{B}^\top (\boldsymbol{B} \boldsymbol{B}^\top)^{-1} \boldsymbol{B}, \label{eq:proj-Bstar} \\
\boldsymbol{A}_{x}^* &= -\boldsymbol{B}^\top (\boldsymbol{B} \boldsymbol{B}^\top)^{-1} \boldsymbol{A}_{x}, \label{eq:proj-Axstar} \\
\boldsymbol{b}^*     &= \boldsymbol{B}^\top (\boldsymbol{B} \boldsymbol{B}^\top)^{-1} \boldsymbol{b}. \label{eq:proj-bstar}
\end{align}
\end{subequations}
\end{ceqn}

\vspace{1 cm}
\noindent \textbf{Special Case 2: Equality constraints with linearity in both \(\boldsymbol{x}\) and \(\boldsymbol{y}\)}

Given a linear constraint $\boldsymbol{A}\hat{\boldsymbol{x}} + \boldsymbol{B}\boldsymbol{y} = \boldsymbol{b}$, the projection problem becomes:

\begin{ceqn}
\begin{equation}
\label{eq:proj_case3}
\tilde{\boldsymbol{y}} = \arg\min_{\boldsymbol{y}} \; \frac{1}{2} \| \boldsymbol{y} - \hat{\boldsymbol{y}}_0 \|^2 \quad \text{s.t.} \quad \boldsymbol{A} \hat{\boldsymbol{x}} + \boldsymbol{B} \boldsymbol{y} = \boldsymbol{b}
\end{equation}
\end{ceqn}

The solution from KKT conditions is:
\begin{ceqn}
\begin{equation}
\label{eq-analytic-linear-x}
\tilde{\boldsymbol{y}} = \hat{\boldsymbol{y}}_0 - \boldsymbol{B}^\top (\boldsymbol{B} \boldsymbol{B}^\top)^{-1}(\boldsymbol{B} \hat{\boldsymbol{y}}_0 + \boldsymbol{A} \hat{\boldsymbol{x}} - \boldsymbol{b})
\end{equation}
\end{ceqn}

This can be written in affine form \cite{chen2024physics}:
\begin{ceqn}
\begin{equation}
\label{eq:lin_affine_final}
\tilde{\boldsymbol{y}} = \boldsymbol{A}^* \hat{\boldsymbol{x}} + \boldsymbol{B}^* \hat{\boldsymbol{y}}_0 + \boldsymbol{b}^*,
\end{equation}
\end{ceqn}
where $\boldsymbol{A}^*$, $\boldsymbol{B}^*$, $\boldsymbol{b}^*$ are given by Eq. \ref{eq:proj-Astar}, \ref{eq:proj-Bstar}, and \ref{eq:proj-bstar} respectively.

To summarize, for constraints involving linear output variables $\boldsymbol{y}$, the KKT conditions allow for an exact, non-iterative projection that can be implemented as a fixed linear layer. However, for general nonlinear constraints defined over both input and output, we resort to Newton-based numerical solution for constraint satisfaction. 

\section{Numerical experiments}
\label{sec:numerical_experiments}
We build all machine learning models using the PyTorch package \cite{paszke2019pytorch}. We use the Adam optimizer \cite{kingma2014adam} for training the models. All numerical experiments are performed on a MacBook Pro with an Apple M4 Pro chip (12-core CPU) and 24 GB of RAM, running macOS. The model implementation and case studies are made available as Jupyter notebook for easy adoption on our GitHub repository at: \url{https://github.com/SOULS-TAMU/kkt-hardnet}.

\paragraph{Performance Metrics.} 
To assess the predictive performance and physical consistency of the models, we report the following metrics: mean squared error (MSE), root mean squared error (RMSE), and mean absolute constraint violation.

For predictions $\hat{\boldsymbol y}_i\in\mathbb{R}^p$ and targets $\bar{\boldsymbol y}_i\in\mathbb{R}^p$,
\[
\mathrm{MSE} \;=\; \frac{1}{N p}\sum_{i=1}^N \sum_{j=1}^p \big(\hat{y}_{ij}-\bar{y}_{ij}\big)^2,
\qquad
\mathrm{RMSE} \;=\; \sqrt{\mathrm{MSE}}.
\]

Mean absolute constraint violation:
\begin{equation}
\label{eq:mean_violation}
\mathrm{Violation}
\;=\;
\frac{1}{N\,m}\sum_{i=1}^N
\left(
\sum_{k\in\mathcal N_E} \big|\,h_k(\boldsymbol x_i,\hat{\boldsymbol y}_i)\,\big|
\;+\;
\sum_{\ell\in\mathcal N_I} \text{ReLU}\big(\,g_\ell(\boldsymbol x_i,\hat{\boldsymbol y}_i)\,\big)
\right),
\qquad
m := |\mathcal N_E|+|\mathcal N_I|,
\end{equation}
\noindent
where $\text{ReLU}(t)=\max\{t,0\}$.  Thus, inequality terms contribute $0$ when $g_\ell(\boldsymbol x_i,\hat{\boldsymbol y}_i)\le 0$ and contribute their magnitude when violated.


\subsection{Example 1: Nonlinearity in both input and output}

We consider the scalar input feature $x\in[1,2]$ and a nonlinear vector-valued target function
\begin{ceqn}
\begin{equation}
\boldsymbol{y}(x)=
\begin{bmatrix}
y_1(x)\\[2pt]
y_2(x)
\end{bmatrix}
=
\begin{bmatrix}
8x^{3}+5\\[2pt]
2x-1
\end{bmatrix},
\qquad
\boldsymbol{y}:\;\mathbb{R}\;\longrightarrow\;\mathbb{R}^2.
\end{equation}
\end{ceqn}

\noindent with \(n_\text{train}=1200\) and \(n_\text{val}=300\). In addition to data \(\{(x_i,\boldsymbol{y}(x_i))\}_{i=1}^n\), we assume knowledge of the following algebraic constraint:

\begin{ceqn}
\begin{equation}
\label{eq:extended_constraint}
h(x,\boldsymbol{y}) := y_1 - y_2^{3} - 12x^{2} + 6x - 6 = 0.
\end{equation}
\end{ceqn}

\noindent The Lagrangian of the following problem becomes:

\begin{ceqn}
\begin{subequations}
\begin{align}
\mathcal{L}(y_{1}, y_{2}, \lambda)
&= \tfrac{1}{2} (y_{1} - \hat{y}_{1, 0})^{2} + \tfrac{1}{2} (y_{2} - \hat{y}_{2, 0})^{2} + \lambda\, h(x, y)
\end{align}
\end{subequations}
\end{ceqn}

The KKT system becomes:

\begin{ceqn}
\begin{subequations}\label{eq:KKT-Case1}
\begin{align}
\frac{\partial \mathcal{L}}{\partial y_1}
&: \; y_1 - \hat{y}_{1,0} + \lambda = 0, \label{eq:ecase1}\\
\frac{\partial \mathcal{L}}{\partial y_2}
&: \; y_2 - \hat{y}_{2,0} - 3\,\lambda\,y_2^2 = 0, \label{eq:ecase2}\\
\frac{\partial \mathcal{L}}{\partial \lambda}
&: \; y_1 - y_2^{3} - 12\,x^{2} + 6\,x - 6 = 0. \label{eq:ecase3}
\end{align}
\end{subequations}
\end{ceqn}

Following the definitions and log-exponential transformations in Section 2.1, the KKT system becomes: 

\begin{ceqn}
\begin{subequations}
\label{eq:full_linexp}
\begin{align}
y_1 - \hat{y}_{1,0} + \lambda &= 0, \label{eq:e1} \\
y_2 - \hat{y}_{2,0} - 3z_{1} &= 0, \label{eq:e2} \\
y_1 - z_{2} - 12x_2 + 6x_1 - 6 &=0, \label{eq:e3} \\
z_{3} + 2 z_{4} - z_{5} &= 0, \label{eq:e4} \\
3 z_{4} - z_{6} &= 0, \label{eq:e5} \\
2 x_{3} - x_{4} = 0, \label{eq:e6} \\
\lambda - e^{z_{3}} = 0, \label{eq:e7} \\
y_{2} - e^{z_{4}} = 0, \label{eq:e8} \\
z_{1} - e^{z_5} = 0, \label{eq:e9} \\
z_{2} - e^{z_{6}} = 0, \label{eq:e10} \\
x_{1} - e^{x_{3}} = 0, \label{eq:e11} \\
x_{2} - e^{x_{4}} = 0 \label{eq:e12}
\end{align}
\end{subequations}
\end{ceqn}

\noindent where, $\hat{y}_{1,0} \text{ and } \hat{y}_{2,0}$ are unconstrained neural net predictions.  

Next, we apply Algorithm \ref{alg:KKT-Hardnet} to seek the closest point on the manifold constructed by equalities in \eqref{eq:full_linexp}. We specify thirty Newton steps (\(K=30\)) with tolerance
\(\lVert\boldsymbol{F}\rVert_\infty<10^{-6}\). The projector
\(
   \rho(\boldsymbol{\hat{y}}_0)=
   ({\tilde{y}}^{(K)}_1,{\tilde{y}}^{(K)}_2)
\)
is fully differentiable, enabling end-to-end training. We compare three different strategies for handling algebraic constraints: (i) an unconstrained multilayer perception (MLP), (ii) a soft constrained PINN, and (iii) KKT-Hardnet. Vanilla MLP involves two hidden layers
\(\bigl[1\!\!\rightarrow\!\!64\!\!\rightarrow\!\!64\!\!\rightarrow\!\!2\bigr]\)
with \(\mathrm{ReLU}\) activation function. KKT-Hardnet consists of the same backbone as the MLP, followed by the projection layer.  For PINN, with an identical backbone to MLP, the loss is augmented by a
soft penalty
\(
   \mathcal L = 
   \text{MSE}
   + \omega (y_1 - y_3 - 12x^{2} + 6x-6)^2
\). All models are trained with Adam (\( \text{lr}=10^{-4} \)) for 1200 epochs. The PINN uses penalty weight \(\omega=100\).

\begin{figure}[t]
   \centering
   \includegraphics[width=1\linewidth]{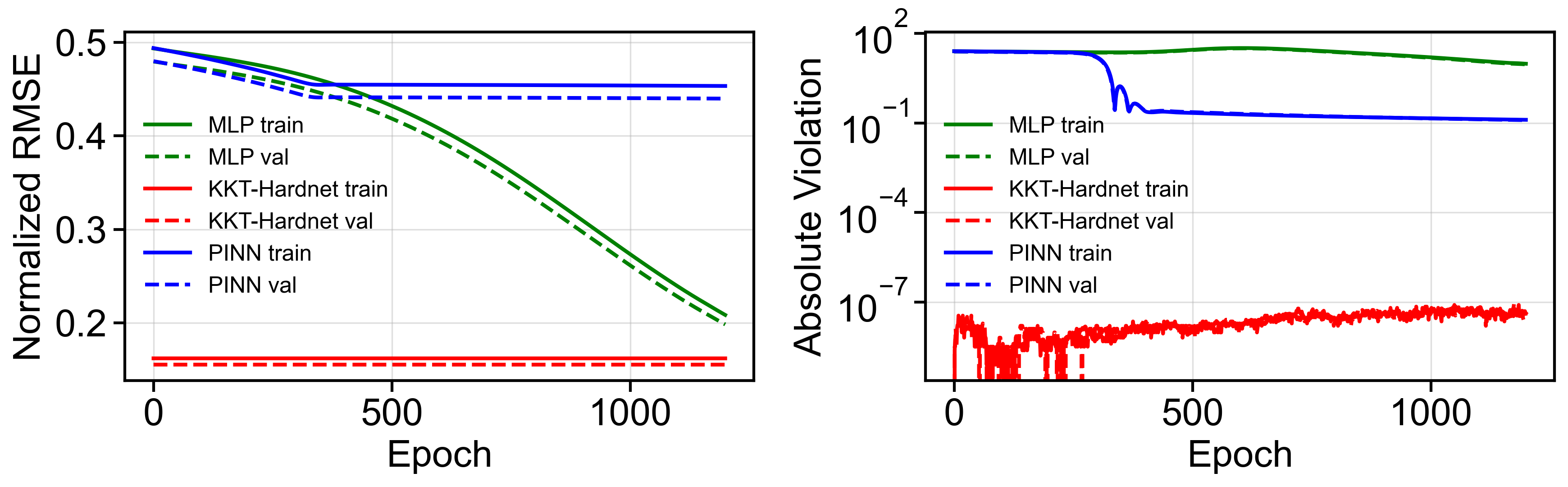}
   \caption{Learning curve for example 1. Left: Normalised RMSE (RMSE is normalized by the range of outputs), Right: constraint violation over 1200 epochs.}
   \label{fig:training_curves}
\end{figure}

Figure \ref{fig:training_curves} shows the learning curves for
normalized RMSE and absolute constraint violation.  The KKT-Hardnet
converges nearly an order of magnitude faster and enforces the
constraint down to numerical precision, whereas the PINN only
reduces the violation to \(\mathcal O(10^{-1})\) despite the
large penalty weight.  Table~\ref{tab:toy_metrics-1} summarizes performance on both training and validation data. KKT-Hardnet reduces constraint violation by 7–8 orders of magnitude while maintaining lower MSE than the unconstrained baseline. In contrast, PINN incurs large MSE errors due to the difficulty of soft constraint enforcement. 


\begin{table}[h]
\centering
\caption{Regression accuracy and constraint violation of KKT-Hardnet compared with an MLP and soft-constrained PINN for example 1.}
\label{tab:toy_metrics-1}
\begin{tabular}{lcccc}
\toprule
& \multicolumn{2}{c}{\textbf{Training}} & \multicolumn{2}{c}{\textbf{Validation}}\\
\cmidrule(lr){2-3}\cmidrule(lr){4-5}
Model   & MSE                   & $|h|$               & MSE                   & $|h|$               \\
\midrule
MLP     & $1.361\times10^{2}$   & $9.61$              & $1.235\times10^{2}$   & $9.13$              \\
PINN    & $6.445\times10^{2}$   & $1.27\times10^{-1}$ & $6.066\times10^{2}$   & $1.24\times10^{-1}$ \\
KKT-Hardnet   & $8.253\times10^{1}$   & $4.21\times10^{-8}$ & $7.585\times10^{1}$   & $3.50\times10^{-8}$ \\
\bottomrule
\end{tabular}
\end{table}

\subsection{Example 2: Nonlinearity in input only}

We consider a two-dimensional input \(\boldsymbol{x} = (x_1, x_2) \in [1,2]^2\), and define a vector-valued ground truth function \(\boldsymbol{y}(\boldsymbol{x}) = (y_1(\boldsymbol{x}), y_2(\boldsymbol{x}))^\top\) as:
\begin{ceqn}
\begin{equation}
\boldsymbol{y}(x_1,x_2) = 
\begin{bmatrix}
x_1^2 + x_2^2\\
4x_1^2 + 4x_2^3 - 2x_2^2
\end{bmatrix},
\qquad
\boldsymbol{y}: \mathbb{R}^2 \to \mathbb{R}^2.
\end{equation}
\end{ceqn}

We assume the following constraint holds:

\begin{ceqn}
\begin{equation}
\label{eq:original_constraint_case2}
y_1 + \tfrac{1}{2}y_2 = 3x_1^2 + 2x_2^3.
\end{equation}
\end{ceqn}

We then introduce auxiliary variables:
\begin{ceqn}
\begin{subequations}
\begin{align}
x_3 &= x_1^2,        & \quad x_4 &= \log x_1,   & \quad x_5 &= \log x_3, \\
x_6 &= x_2^3,        & \quad x_7 &= \log x_2,   & \quad x_8 &= \log x_6.
\end{align}
\end{subequations}
\end{ceqn}

This gives the following augmented system:
\begin{ceqn}
\begin{subequations}
\label{eq:case2_expanded_constraints}
\begin{align}
y_1 + \tfrac12\,y_2 - 3x_3 - 2x_6 &= 0, \label{eq:c1_case2}\\
2 x_{4} - x_{5} &= 0, \label{eq:c2_case2}\\
3 x_{7} = x_{8} &= 0, \label{eq:c3_case2}\\
x_{1} - e^{x_{4}} &= 0, \label{eq:c4_case2}\\
x_{3} - e^{x_{5}} &= 0, \label{eq:c5_case2}\\
x_{2} - e^{x_{7}} &= 0, \label{eq:c6_case2}\\
x_{6} - e^{x_{8}} &= 0. \label{eq:c7_case2}
\end{align}
\end{subequations}
\end{ceqn}

We collect all eight inputs as \(\boldsymbol{x} = (x_1,\dots,x_8)^\top \in \mathbb{R}^8\), and express the constraint system in the general form (see Eq. \ref{eq:log_e1}) as follows:

\[
\boldsymbol{A} = 
\begin{bmatrix}
 0 & 0 & -3 & 0 &  0 &  -2 &  0 &  0\\
 0 & 0 &  0 &  2 &  -1 & 0 &  0 &  0\\
 0 & 0 &  0 &  0 &  0 &  0 &  3 &  -1\\
 1 & 0 &  0 &  0 &  0 &  0 &  0 &  0\\
 0 & 0 &  1 &  0 &  0 &  0 &  0 & 0\\
 0 & 1 &  0 &  0 &  0 &  0 &  0 &  0\\
 0 & 0 &  0 &  0 &  0 &  1 &  0 &  0
\end{bmatrix},\quad
\boldsymbol{B} = 
\begin{bmatrix}
 1 & \tfrac12\\
 0 & 0\\
 0 & 0\\
 0 & 0\\
 0 & 0\\
 0 & 0\\
 0 & 0
\end{bmatrix},\quad
\boldsymbol{A_{x}} =
\begin{bmatrix}
 0 & 0 & 0 & 0 & 0 & 0 & 0 & 0\\
 0 & 0 & 0 & 0 & 0 & 0 & 0 & 0\\
 0 & 0 & 0 & 0 & 0 & 0 & 0 & 0\\
 0 & 0 & 0 & -1 & 0 & 0 & 0 & 0\\
 0 & 0 & 0 & 0 & -1 & 0 & 0 & 0\\
 0 & 0 & 0 & 0 & 0 & 0 & -1 & 0\\
 0 & 0 & 0 & 0 & 0 & 0 & 0 & -1
\end{bmatrix}.
\]

\vspace{1 cm}
\textbf{Analytic Projection:} We notice that, only the first row of $\boldsymbol{B}$ depends on the output, corresponding to:

\[
\boldsymbol{B}_{\text{active}} \boldsymbol{y} = 3x_3 + 2x_6,
\quad\text{with}\quad
\boldsymbol{B}_{\text{active}} = \begin{bmatrix} 1 & \tfrac12 \end{bmatrix}.
\]

Rows 2 to 7 do not depend on $\boldsymbol{y}$. Therefore, for any given input $x_1, x_2$, all terms in rows 2 to 7 are already satisfied, and thus, they do not participate in the projection. The projection of an unconstrained neural net prediction \(\boldsymbol{\hat{y}}_0\) onto this hyperplane is therefore, given by the analytic form:
\begin{ceqn}
\begin{equation}
\label{eq:analytic_proj_case2}
\tilde{\boldsymbol{y}} = \boldsymbol{\hat{y}}_0
- \boldsymbol{B}_{\text{active}}^\top
\left(\boldsymbol{B}_{\text{active}} \boldsymbol{B}_{\text{active}}^\top\right)^{-1}
\left(\boldsymbol{B}_{\text{active}} \boldsymbol{\hat{y}}_0 - 3x_3 - 2x_6\right).
\end{equation}
\end{ceqn}

Since \(\boldsymbol{B}_{\text{active}} \boldsymbol{B}_{\text{active}}^\top = \tfrac54\), the scalar residual $r$ can be defined as:
\[
r = \left(\boldsymbol{B}_{\text{active}} \boldsymbol{\hat{y}}_0 - 3x_3 - 2x_6\right) =  \hat{y}_{0,1} + 0.5\,\hat{y}_{0,2} - 3x_3 - 2x_6,
\qquad\text{and}\qquad
\left(\boldsymbol{B}_{\text{active}} \boldsymbol{B}_{\text{active}}^\top\right)^{-1} = \tfrac{4}{5}.
\]
The final projection update becomes:

\begin{ceqn}
\begin{equation}
\label{eq:analytic_proj_case2}
\tilde{\boldsymbol{y}} = \boldsymbol{\hat{y}}_0
- \frac{4}{5} \begin{bmatrix} 1 \\ \tfrac12 \end{bmatrix}r .
\end{equation}
\end{ceqn}

\begin{ceqn}
\begin{equation}
\tilde{y}_1 = \hat{y}_{0,1} - 0.8\,r,
\qquad
\tilde{y}_2 = \hat{y}_{0,2} - 0.4\,r.
\label{eq:analytic-form}
\end{equation}    
\end{ceqn}

Note that, if one wishes to consider the entire $\boldsymbol{B} \in \mathbb{R}^{7\times2}$, the analytic projection does not change, except $\boldsymbol{B}\boldsymbol{B}^\top$ becomes singular. In that case, one needs to replace the regular inverse with a Moore–Penrose pseudo‐inverse \cite{barata2012moore}. The general projection formula collapses exactly to the same update as given in Eq. \ref{eq:analytic-form}, because all the zero rows in $\boldsymbol{B}$ drop out of the pseudoinverse calculation. 

\begin{figure}[t]
   \centering
   \includegraphics[width=1\linewidth]{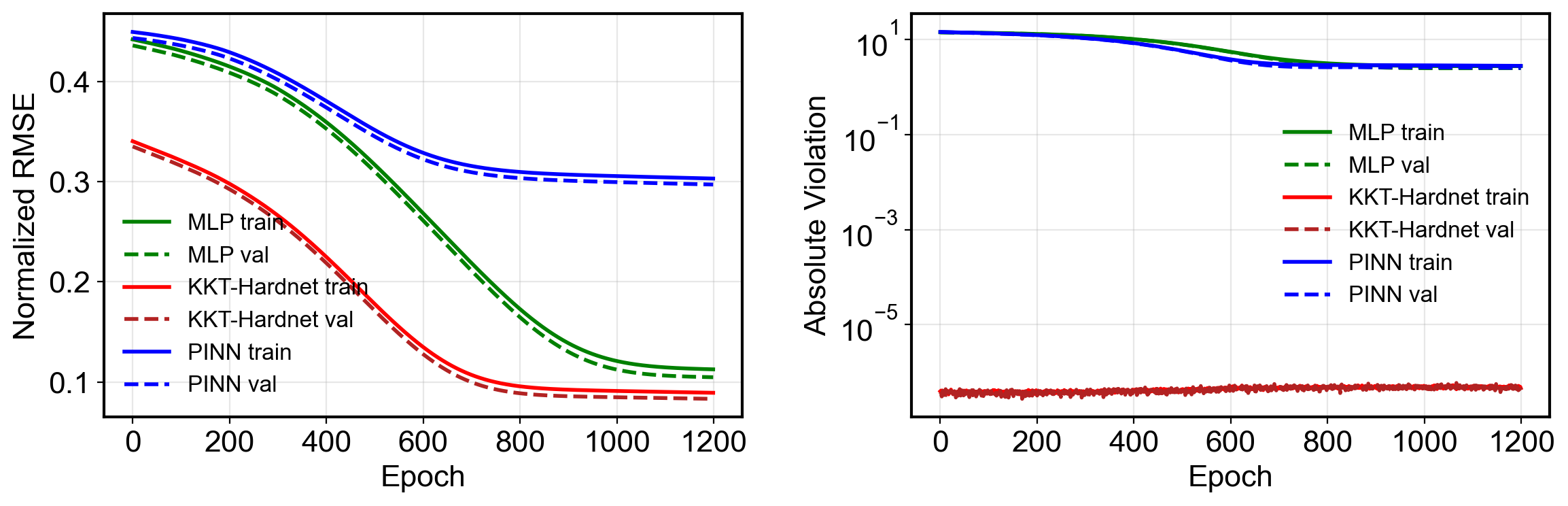}
   \caption{Learning curves for example 2. Left: Normalized RMSE; Right: absolute constraint violation over 1200 epochs.}
   \label{fig:training_curves_case2}
\end{figure}

\textbf{Training results:} We train three models using identical MLP architectures (2-64–64–2) with ReLU activations and the Adam optimizer (\(\text{lr}=10^{-4}\), 1200 epochs). Model 1 is MLP which is an unconstrained neural network trained via MSE loss, Model 2 is PINN where MSE loss with soft penalty \(\omega (y_1 + 0.5y_2 - (3x_1^2 + 2x_2^3))\), \(\omega=100\) is applied. Finally, Model 3 is KKT-Hardnet where analytic projection, i.e., Eq. \ref{eq:analytic-form}, is applied after raw prediction from the MLP backbone. As shown in Figure \ref{fig:training_curves_case2} and Table \ref{tab:toy_metrics_case2}, KKT-Hardnet reduces constraint violation by over 6 orders of magnitude while achieving the lowest validation error. In contrast, the soft-penalty PINN model suffers from large errors due to inadequate constraint enforcement.

\begin{table}[h]
\centering
\caption{Regression accuracy and constraint enforcement comparison for the illustrative example 2.}
\label{tab:toy_metrics_case2}
\begin{tabular}{lcccc}
\toprule
& \multicolumn{2}{c}{\textbf{Training}} & \multicolumn{2}{c}{\textbf{Validation}}\\
\cmidrule(lr){2-3}\cmidrule(lr){4-5}
Model & MSE & $|h|$ & MSE & $|h|$ \\
\midrule
MLP & \(1.443\!\times\!10^{1}\) & 2.77 & \(1.248\!\times\!10^{1}\) & 2.51 \\
PINN & \(1.045\!\times\!10^{2}\) & 2.81 & \(1.004\!\times\!10^{2}\) & 2.55 \\
KKT-Hardnet & \(9.051\!\times\!10^{0}\) & \(4.60\!\times\!10^{-7}\) & \(7.863\!\times\!10^{0}\) & \(4.23\!\times\!10^{-7}\) \\
\bottomrule
\end{tabular}
\end{table}

\subsection{Example 3: Inequality constraints}
\label{sec:ineq_example}

We demonstrate the hard enforcement of inequality with the task of learning $y = x^2$ subject to the scalar inequality $y - x \;\le\; 0$. We generate data by uniformly sampling $x$ such that, 

\begin{ceqn}
\begin{equation}
x_i \sim \mathcal{U}(1, 2), \qquad y_i = x_i^2,
\qquad
\begin{cases}
i = 1,\dots,1200 & \text{(training data)} \\
i = 1201,\dots,1500 & \text{(validation data)}
\end{cases}
\end{equation}
\end{ceqn}

\noindent We fit a vanilla MLP  
\(\hat y_0 = \mathrm{NN}_{\boldsymbol{\Theta}}(x)\in\mathbb R\)  
and then project \(\hat y_0\) onto \(\{y\mid y\le x\}\) by performing the following projection:

\begin{ceqn}
\begin{align}
\min_{y} \quad & \tfrac{1}{2}(y - \hat{y}_0)^2 \\
\text{s.t.} \quad & y - x + s = 0, \\
& s \ge 0
\end{align}
\end{ceqn}

The KKT system is then derived as follows:

\begin{ceqn}
  \begin{align}
y-\hat{y}_0+\mu^E &=0,
  \label{eq:stat_general-1}\\[4pt]
\mu^E - \mu^I &=0,
  \label{eq:prim_general}\\[4pt]
y - x + s &=0,
  \label{eq:dual_general}\\[4pt]
\mu^I\,s &= 0,
  \label{eq:comp_general_1} \\
  \mu^I &\geq 0, \\
  s &\geq 0
\end{align}
\end{ceqn}

Next, we define the following auxiliary variables for the log-exp reformulation:

\begin{ceqn}
\begin{subequations}
\begin{align}
z_3     &= \log(\mu),          & \quad z_5     &= \log(s),          \label{eq:w3w5} \\
z_1     &= {\mu^I}^2,              & \quad z_2     &= s^2,              \label{eq:w1w2} \\
z_7     &= z_1 + z_2,          & \quad z_4     &= 2\,z_3,           \label{eq:w7w4} \\
z_6     &= 2\,z_5,             & \quad z_{10}  &= \log(z_7),        \label{eq:w6w10} \\
z_9     &= \tfrac{1}{2}\,z_{10}, & \quad z_8     &= e^{z_9}.         \label{eq:w9w8}
\end{align}
\end{subequations}
\end{ceqn}

These auxiliary variables during log-exp reformulation ensure
that at initialization, all logarithmic and exponential relations hold
exactly, and that \(\mu^I>0,\,s>0\) throughout the Newton iterations. At the start of each forward pass we set

\begin{ceqn}
\begin{subequations}
\begin{align}
y       &= \hat{y}_0, \qquad
\mu^E = 0, \label{eq:ystationary} \\
\mu^I     &= \mathrm{ReLU}(\hat{y}_0 - x) + \epsilon, \label{eq:mu-def} \\
s       &= \mathrm{ReLU}(x - \hat{y}_0) + \epsilon. \label{eq:s-def}
\end{align}
\end{subequations}
\end{ceqn}

with \(\epsilon=10^{-3}\).

After the reformulation, the KKT system becomes

\begin{ceqn}
\begin{subequations}\label{eq:kkt_ineq}
\begin{align}
F_{1}(\boldsymbol{\tau})  &:~ y - \hat{y}_0 + \mu^E = 0, \label{eq:F1}\\
F_{2}(\boldsymbol{\tau})  &:~ \mu^E - \mu^I = 0, \label{eq:F2}\\
F_{3}(\boldsymbol{\tau})  &:~ y - x + s = 0, \label{eq:F3}\\
F_{4}(\boldsymbol{\tau})  &:~ z_{8} - \mu^I - s = 0, \label{eq:F4}\\
F_{5}(\boldsymbol{\tau})  &:~ \mu^I - e^{z_{3}} = 0, \label{eq:F5}\\
F_{6}(\boldsymbol{\tau})  &:~ z_{1} - e^{z_{4}} = 0, \label{eq:F6}\\
F_{7}(\boldsymbol{\tau})  &:~ 2\,z_{3} - z_{4} = 0, \label{eq:F7}\\
F_{8}(\boldsymbol{\tau})  &:~ s - e^{z_{5}} = 0, \label{eq:F8}\\
F_{9}(\boldsymbol{\tau})  &:~ z_{2} - e^{z_{6}} = 0, \label{eq:F9}\\
F_{10}(\boldsymbol{\tau}) &:~ 2\,z_{5} - z_{6} = 0, \label{eq:F10}\\
F_{11}(\boldsymbol{\tau}) &:~ z_{7} - z_{1} - z_{2} = 0, \label{eq:F11}\\
F_{12}(\boldsymbol{\tau}) &:~ z_{9} - \tfrac{1}{2}\,z_{10} = 0, \label{eq:F12}\\
F_{13}(\boldsymbol{\tau}) &:~ z_{8} - e^{z_{9}} = 0, \label{eq:F13}\\
F_{14}(\boldsymbol{\tau}) &:~ z_{7} - e^{z_{10}} = 0. \label{eq:F14}
\end{align}
\end{subequations}
\end{ceqn}

\noindent where all the variables are compactly represented by a vector $\boldsymbol{\tau} = [\boldsymbol{y}, \boldsymbol{z}, \boldsymbol{\lambda}]$ as follows:  
\[
  \boldsymbol{\tau} = \bigl[y,\,s,\,\mu^I,\,\mu^E,\,
          z_3,\,z_5,\,
          z_1,\,z_2,\,z_7,\,
          z_4,\,z_6,\,
          z_8,\,z_9,\,z_{10}
      \bigr]^\top
\]
These residuals are solved by a damped Gauss–Newton iteration to specified tolerance, yielding the projected output \(\tilde y\) (see Algorithm \ref{alg:KKT-Hardnet}).

\begin{figure}[t]
   \centering
   \includegraphics[width=1\linewidth]{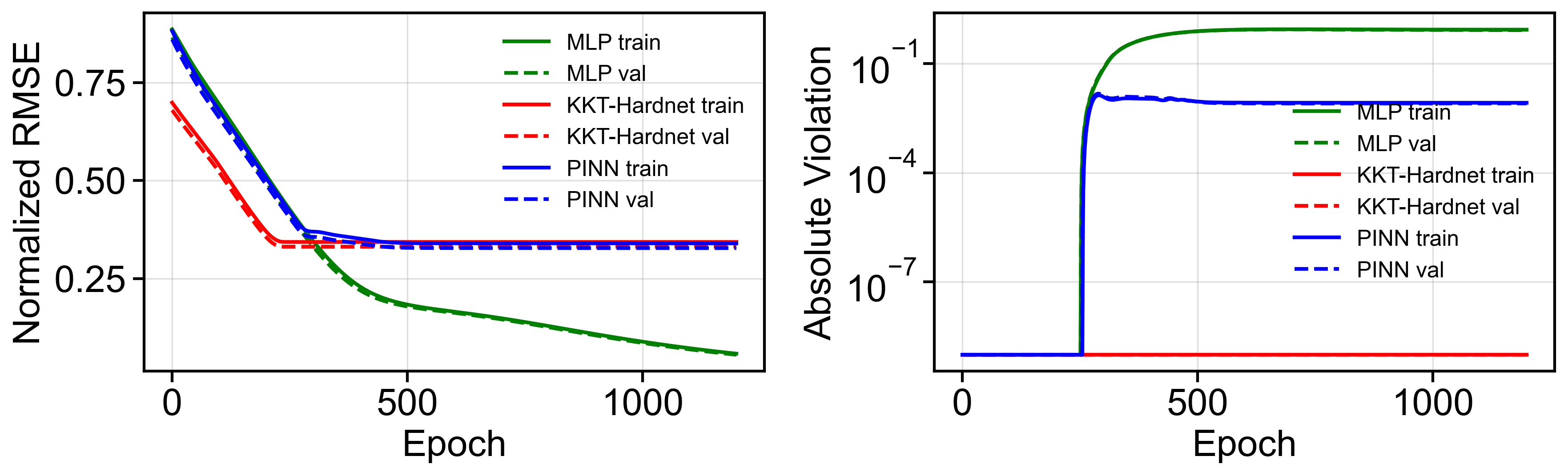}
   \caption{Learning curves for example 3. Left: Normalized RMSE; Right: absolute constraint violation (log scale).}
   \label{fig:inequality}
\end{figure}

\textbf{Model training and comparison.} 
We compare KKT-Hardnet with an unconstrained MLP and a soft-constrained PINN, all using identical architectures and training routines. Figure~\ref{fig:inequality} shows the normalized RMSE and absolute constraint violation for all three models. The results in Table~\ref{tab:inequality} clearly show that KKT-Hardnet maintains constraint satisfaction up to numerical precision while achieving comparable MSE to the unconstrained MLP. In contrast, the PINN suffers from persistent constraint violations despite soft penalty enforcement.

\begin{table}[H]
\centering
\caption{Regression accuracy and constraint enforcement comparison for the illustrative example 3.}
\label{tab:inequality}
\begin{tabular}{lcccc}
\toprule
& \multicolumn{2}{c}{\textbf{Training}} & \multicolumn{2}{c}{\textbf{Validation}}\\
\cmidrule(lr){2-3}\cmidrule(lr){4-5}
Model   & MSE                     & \(|g|\)              & MSE                     & \(|g|\)              \\
\midrule
MLP           & \(3.038\!\times\!10^{-2}\) & \(8.48\!\times\!10^{-1}\) & \(2.767\!\times\!10^{-2}\) & \(8.22\!\times\!10^{-1}\) \\
PINN          & \(1.037\!\times\!10^{0}\)  & \(8.32\!\times\!10^{-3}\) & \(9.648\!\times\!10^{-1}\) & \(8.00\!\times\!10^{-3}\) \\
KKT-Hardnet   & \(1.059\!\times\!10^{0}\)  & \(1.00\!\times\!10^{-9}\) & \(9.857\!\times\!10^{-1}\) & \(1.00\!\times\!10^{-9}\) \\
\bottomrule
\end{tabular}
\end{table}

\subsection{A Case Study on Chemical Process Simulation Involving Nonlinear Thermodynamics}

\label{sec:ED-column}

\begin{figure}[t]
   \centering
   \includegraphics[scale = 0.8]{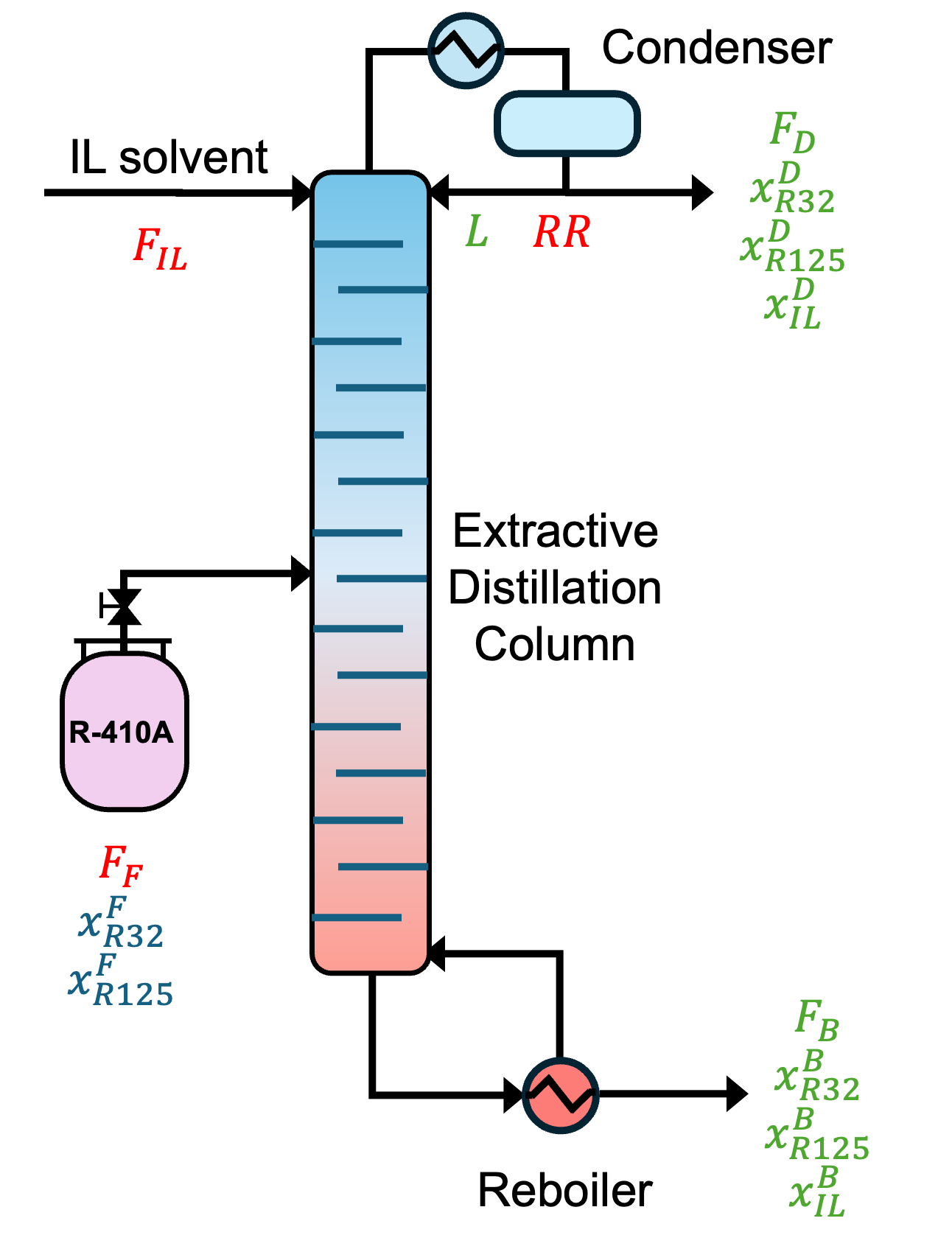}
   \caption{Extractive distillation of R-410A using an Ionic liquid [EMIM][SCN]. The input variables are denoted in red. The output variables are denoted in green. The fixed parameters are denoted in blue. The column operates at a fixed pressure of 10 bar.}
   \label{fig:flowsheet}
\end{figure}

We consider the simulation of an extractive distillation process for separating the azeotropic refrigerant mixture R-410A using the ionic liquid \([ \mathrm{EMIM} ][ \mathrm{SCN} ]\). The flowsheet is given in Figure \ref{fig:flowsheet}, which consists of a distillation column with an ionic liquid entrainer fed at a fixed column stage. The feed R-410A is an azeotropic mixture of R-32 and R-125. Therefore, the mass composition of R-32 and R-125 in R-410A feed are 0.5 and 0.5, respectively. [EMIM][SCN] and R-410A are assumed to be fed at stage 2 and 11 respectively. The total number of stages of the columns is set to 18. Due to the presence of ionic liquids as entrainer, highly nonlinear thermodynamics is present for the calculation of phase equilibria across the extractive distillation column. In this case study, all simulations are performed using equilibrium-stage modeling using NRTL model in Aspen Plus\textsuperscript{\textregistered} v12. Details can be found elsewhere \cite{iftakher2025multi, iftakher2025integrating, monjur2022separation}.

\subsubsection{Problem setup and data generation}
We vary three input variables: total feed molar flow rate \((F_{\mathrm{F}})\), ionic liquid flow rate \((F_{\mathrm{IL}})\), and the reflux ratio \((R)\), and extract a set of steady-state outputs from Aspen Plus. The input space is sampled on a uniform grid: (1) Feed flow rate: \(F_{\mathrm{F}} \in [95, 110]\;\mathrm{kg/hr}\), (2) IL flow rate: \(F_{\mathrm{IL}} \in [800, 950]\;\mathrm{kg/hr}\), and (3) Reflux ratio: \(R \in [2.5, 4.3]\). 
The corresponding outputs include: (1) Distillate and bottoms flowrates: \(F_{\mathrm{D}}, F_{\mathrm{B}} \;\;[\mathrm{mol/s}]\), (2) Tray liquid flowrate: \(L \;\;[\mathrm{mol/s}]\), (3) Mole fractions in distillate and bottoms:  $x^{D}_{R32}, x^{D}_{R125}, x^{D}_{IL}, x^{B}_{R32}, x^{B}_{R125}, x^{B}_{IL}$. Not all simulations converged without error. Therefore, we postprocessed all simulations, and filtered out unconverged ones. This resulted in 4308 simulation data points, which constitute our dataset.

\subsubsection{Nonlinear constraints in both input and output}
\label{Sec:nonlinear-ED}
We consider six physical constraints, derived from mass and energy balances involving mixed linearity and nonlinearities in both inputs and outputs:

\begin{enumerate}[label=(C\arabic*), leftmargin=2.2cm]
    \item Overall molar balance (affine):
    \begin{ceqn}
    \begin{equation}
    F_{\mathrm{F}} + F_{\mathrm{IL}} = F_{\mathrm{D}} + F_{\mathrm{B}}
    \end{equation}
    \end{ceqn}

    \item Component balance – R32 (nonlinear in input-output): 
    \begin{ceqn}
    \begin{equation}
    F_{\mathrm{F}}\,x^{\mathrm{F}}_{\mathrm{R32}} = F_{\mathrm{D}}\,x^{\mathrm{D}}_{\mathrm{R32}} + F_{\mathrm{B}}\,x^{\mathrm{B}}_{\mathrm{R32}}
    \end{equation}
    \end{ceqn}

    \item Component balance – R125 (nonlinear in input-output):
    \begin{ceqn}
    \begin{equation}
    F_{\mathrm{F}}\,x^{\mathrm{F}}_{\mathrm{R125}} = F_{\mathrm{D}}\,x^{\mathrm{D}}_{\mathrm{R125}} + F_{\mathrm{B}}\,x^{\mathrm{B}}_{\mathrm{R125}}
    \end{equation}
    \end{ceqn}

    \item Mole fraction balance – Distillate (affine):
    \begin{ceqn}
    \begin{equation}
    x^{\mathrm{D}}_{\mathrm{R32}} + x^{\mathrm{D}}_{\mathrm{R125}} + x^{\mathrm{D}}_{\mathrm{IL}} = 1
    \end{equation}
    \end{ceqn}

    \item Mole fraction closure – Bottoms (affine):
    \begin{ceqn}
    \begin{equation}
    x^{\mathrm{B}}_{\mathrm{R32}} + x^{\mathrm{B}}_{\mathrm{R125}} + x^{\mathrm{B}}_{\mathrm{IL}} = 1
    \end{equation}
    \end{ceqn}

    \item Reflux ratio relation (bilinear in input and output):
    \begin{ceqn}
    \begin{equation}
    R = \frac{L}{F_{\mathrm{D}}}
    \end{equation}
    \end{ceqn}
\end{enumerate}

Here, $x_{R32}^F=0.697616946$ and $x_{R125}^F = 0.302383054$.

\vspace{1 cm}

Next, we enforce the six nonlinear algebraic constraints by constructing the KKT system and log-exponential transformation (see Section \ref{subsec:log_transformation} for details). Specifically, bilinear terms like \( F_{\mathrm{D}} \cdot x^{\mathrm{D}}_{\mathrm{R32}} \) are log transformed and replaced by auxiliary variables, allowing the constraint system to be rewritten in the form given in Eq. \ref{eq:Primal-feasibility_eq}. Then we train KKT-Hardnet (see Algorithm \ref{alg:KKT-Hardnet}. In our implementation, we use \( K = 30 \), \( \gamma = 10^{-3} \), and \( \alpha \in (0,1]\) is adjusted based on backtracking and Armijo condition. The projection operation is fully differentiable and acts as a corrector layer: $\rho(\hat{\boldsymbol{y}}_0) = \boldsymbol{y}^{(K)}$. 

\begin{figure}[t]
   \centering
   \includegraphics[width=1\linewidth]{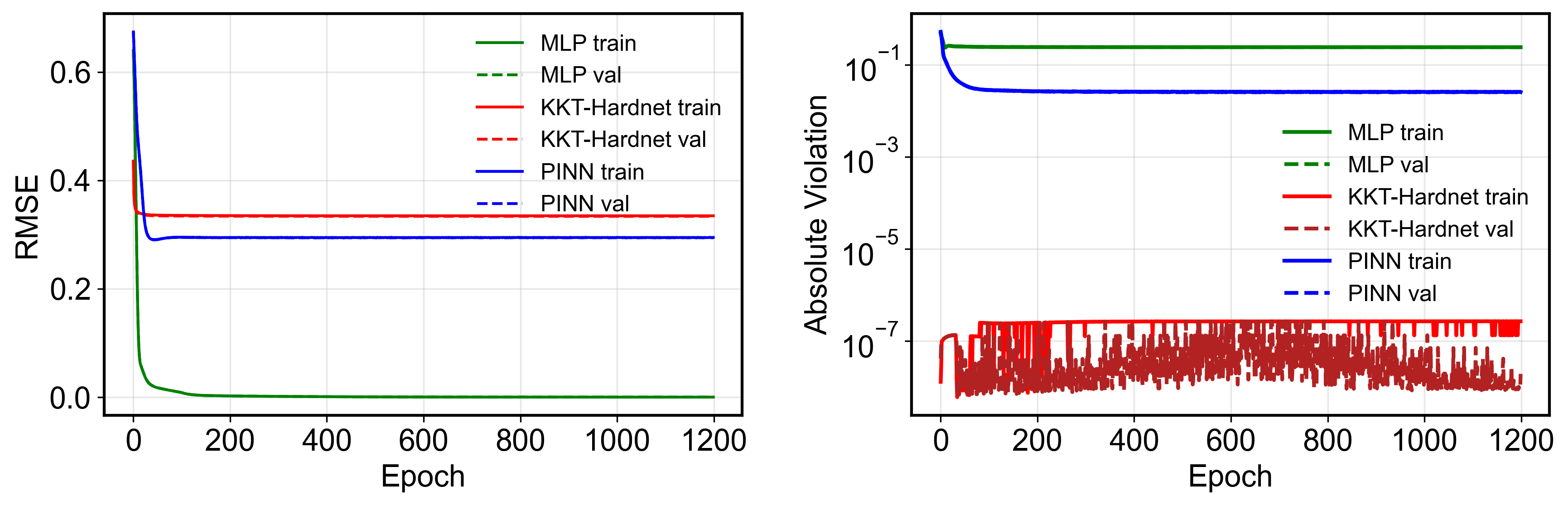}
   \caption{Learning curves for the simulation of extractive distillation-based simulation of R-410A using an ionic liquid [EMIM][SCN]. Left: RMSE; Right: absolute constraint violation over 1200 epochs.}
   \label{fig:training_curves_case3}
\end{figure}

We compare the performance of KKT-Hardnet with a soft penalty-based PINN and an unconstrained MLP, all using identical architectures and optimization settings (learning rate \(10^{-4}\), 1200 epochs, Adam optimizer). The PINN model includes the six physical constraints as penalty terms in the loss function with a penalty weight \(\omega = 10.0\). Figure \ref{fig:training_curves_case3} displays the learning curves for RMSE and absolute constraint violation. We observe no violation spikes above $10^{-6}$, confirming that both training and validation outputs satisfy the constraints to the specified tolerance. The small oscillations below $10^{-6}$ arise because, in some forward passes, the Gauss–Newton iteration terminates with the residual norm dropping well below the stopping criterion of $10^{-6}$. In those cases, the final residual may lie anywhere between the tolerance ($10^{-6}$) and the numerical precision limit. Thus, the fluctuations reflect solver termination under finite precision.


\begin{table}[H]
\centering
\caption{Model performance comparison on the extractive distillation case study.}
\label{tab:distillation_metrics}
\begin{tabular}{lcccc}
\toprule
& \multicolumn{2}{c}{\textbf{Training}} & \multicolumn{2}{c}{\textbf{Validation}}\\
\cmidrule(lr){2-3}\cmidrule(lr){4-5}
Model & MSE & \(|h|\) & MSE & \(|h|\) \\
\midrule
MLP & \(3.146\!\times\!10^{-8}\) & \(2.39\!\times\!10^{-1}\) & \(3.798\!\times\!10^{-8}\) & \(2.39\!\times\!10^{-1}\) \\
PINN & \(8.680\!\times\!10^{-2}\) & \(2.55\!\times\!10^{-2}\) & \(8.662\!\times\!10^{-2}\) & \(2.52\!\times\!10^{-2}\) \\
KKT-Hardnet & \(1.119\!\times\!10^{-1}\) & \(2.70\!\times\!10^{-7}\) & \(1.113\!\times\!10^{-1}\) & \(1.95\!\times\!10^{-8}\) \\
\bottomrule
\end{tabular}
\end{table}

As shown in Table~\ref{tab:distillation_metrics}, the KKT-Hardnet achieves constraint satisfaction up to the specified tolerance (\(\sim 10^{-6}\)). Although its MSE is higher than the unconstrained MLP, it eliminates physical inconsistency and constraint violation. In contrast, the PINN, despite a soft penalty, fails to sufficiently reduce constraint violation, indicating that soft regularization alone is insufficient for strict feasibility in this nonlinear, process-driven setting. An analysis of the PINN penalty weight $\omega$ on MSE and constraint violation is given in Appendix A.

These results support the following observation: MLP tries to minimize the loss between the prediction and the raw data (even if the raw data may not satisfy known physics). On the other hand, KKT-Hardnet tries to minimize the difference between a physics-satisfying output and the raw data. Therefore, it may be that the RMSE for KKT-Hardnet is larger than MLP. However, the constraint violation is eliminated. PINN, on the other hand, offers a tradeoff (higher loss than MLP but lower violation than MLP).

\subsubsection{Analytic projection for affine-in-output constraints}
\label{sec:linear-ED}
We consider a special case which consists of a subset of four output constraints that are linear in outputs. These include the overall molar balance, the mole fraction balances for distillate and bottoms, and the reflux relation. Specifically:

\begin{enumerate}[label=(C\arabic*), leftmargin=2.2cm]
    \item[(C1)] Overall molar balance:
    \begin{ceqn}
    \begin{equation}
    F_{\mathrm{F}} + F_{\mathrm{IL}} = F_{\mathrm{D}} + F_{\mathrm{B}}
    \end{equation}
    \end{ceqn}

    \item[(C4)] Mole fraction closure — Distillate:
    \begin{ceqn}
    \begin{equation}
    x^{\mathrm{D}}_{\mathrm{R32}} + x^{\mathrm{D}}_{\mathrm{R125}} + x^{\mathrm{D}}_{\mathrm{IL}} = 1
    \end{equation}
    \end{ceqn}

    \item[(C5)] Mole fraction closure — Bottoms:
    \begin{ceqn}
    \begin{equation}
    x^{\mathrm{B}}_{\mathrm{R32}} + x^{\mathrm{B}}_{\mathrm{R125}} + x^{\mathrm{B}}_{\mathrm{IL}} = 1
    \end{equation}
    \end{ceqn}

    \item[(C6)] Reflux ratio relation:
    \begin{ceqn}
    \begin{equation}
    L = R \cdot F_{\mathrm{D}}
    \end{equation}
    \end{ceqn}
\end{enumerate}

These constraints define a linear manifold over the 9-dimensional output space. Given a raw neural network prediction \( \hat{\boldsymbol{y}}_0 \in \mathbb{R}^{9} \), we project it onto the feasible manifold using an analytic projection given in Eq. \ref{Eq-Analytic-Nonlinear-x}:

\begin{ceqn}
\begin{equation}
\boldsymbol{y}_{\text{proj}} = \hat{\boldsymbol{y}}_0 - \boldsymbol{B}^\top(\boldsymbol{B}\boldsymbol{B}^\top)^{-1}\boldsymbol{r},
\end{equation}
\end{ceqn}

where \( \boldsymbol{B} \in \mathbb{R}^{4 \times 9} \) encodes the output coefficients of the constraints, and \( \boldsymbol{r} \in \mathbb{R}^{4} \) is the residual vector formed from the violation of each constraint. Specifically, for a given input \( \boldsymbol{x} = (F_{\mathrm{F}}, F_{\mathrm{IL}}, R) \), the residual vector is computed as:

\begin{ceqn}
\begin{equation}
\boldsymbol{r} = 
\begin{bmatrix}
F_{\mathrm{F}} + F_{\mathrm{IL}} - (\hat{F}_{\mathrm{D}} + \hat{F}_{\mathrm{B}})\\
\hat{x}^{\mathrm{D}}_{\mathrm{R32}} + \hat{x}^{\mathrm{D}}_{\mathrm{R125}} + \hat{x}^{\mathrm{D}}_{\mathrm{IL}} - 1\\
\hat{x}^{\mathrm{B}}_{\mathrm{R32}} + \hat{x}^{\mathrm{B}}_{\mathrm{R125}} + \hat{x}^{\mathrm{B}}_{\mathrm{IL}} - 1\\
- R \cdot \hat{F}_{\mathrm{D}} + \hat{L}
\end{bmatrix}
\end{equation}
\end{ceqn}

This projection is performed independently for each sample in the batch, and requires no iterative solver or Jacobian computation. The matrix \( \boldsymbol{B} \) is fixed for each sample based on known input \( \boldsymbol{x} \). As such, the projection step is differentiable and computationally efficient.
 
\begin{figure}[t]
   \centering
   \includegraphics[width=1\linewidth]{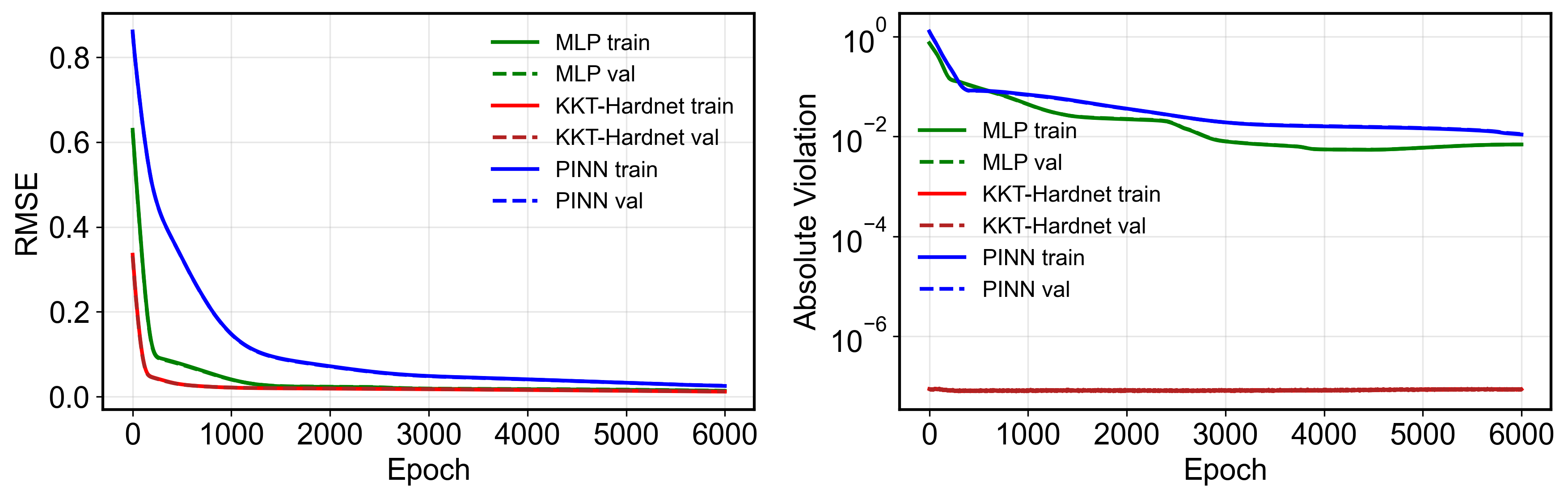}
   \caption{Learning curves for the analytic projection case. Left: RMSE; Right: absolute constraint violation (log scale).}
   \label{fig:analytic_proj_curves}
\end{figure}

We compare KKT-Hardnet with an analytic projection layer against an unconstrained MLP and a soft-constrained PINN, all using identical architectures and training routines. Figure~\ref{fig:analytic_proj_curves} shows the RMSE and constraint violation for all three models. The results in Table~\ref{tab:analytic_proj_metrics} clearly show that the analytically projected KKT-Hardnet maintains constraint satisfaction up to numerical precision while achieving lower MSE than the unconstrained MLP. In contrast, the PINN suffers from persistent constraint violations despite soft penalty enforcement. An analysis of the PINN penalty weight $\omega$ on MSE and constraint violation is given in Appendix A.

\begin{table}[H]
\centering
\caption{Model performance for affine-in-output constraints for the extractive distillation case study.}
\label{tab:analytic_proj_metrics}
\begin{tabular}{lcccc}
\toprule
& \multicolumn{2}{c}{\textbf{Training}} & \multicolumn{2}{c}{\textbf{Validation}}\\
\cmidrule(lr){2-3}\cmidrule(lr){4-5}
Model & MSE & \(|h|\) & MSE & \(|h|\) \\
\midrule
MLP  & \(1.828\!\times\!10^{-4}\) & \(6.89\!\times\!10^{-3}\) &
       \(1.932\!\times\!10^{-4}\) & \(6.90\!\times\!10^{-3}\) \\
PINN & \(6.359\!\times\!10^{-4}\) & \(1.09\!\times\!10^{-2}\) &
       \(6.505\!\times\!10^{-4}\) & \(1.11\!\times\!10^{-2}\) \\
KKT-Hardnet & \(1.390\!\times\!10^{-4}\) & \(8.66\!\times\!10^{-8}\) &
       \(1.479\!\times\!10^{-4}\) & \(8.61\!\times\!10^{-8}\) \\
\bottomrule
\end{tabular}
\end{table}

This case highlights that when constraints are affine or linear in outputs, an exact and efficient analytic projection can be used to enforce strict feasibility without iterative solvers. In our dataset, the Aspen Plus simulations already satisfy these four constraints to near machine precision (maximum residuals $\approx 10^{-9}$). In such a case, KKT-Hardnet offers much faster convergence than MLP. It also eliminates constraint violation. This illustrates that KKT-Hardnet projected outputs (instead of raw MLP outputs) guide the backpropagation in a way such that the loss function converges faster than the MLP.

\subsection{A Case Study on Pooling Problem Involving Nonlinear Equality and Inequality Constraints}
\begin{figure}[t]
   \centering
   \includegraphics[width=0.6\linewidth]{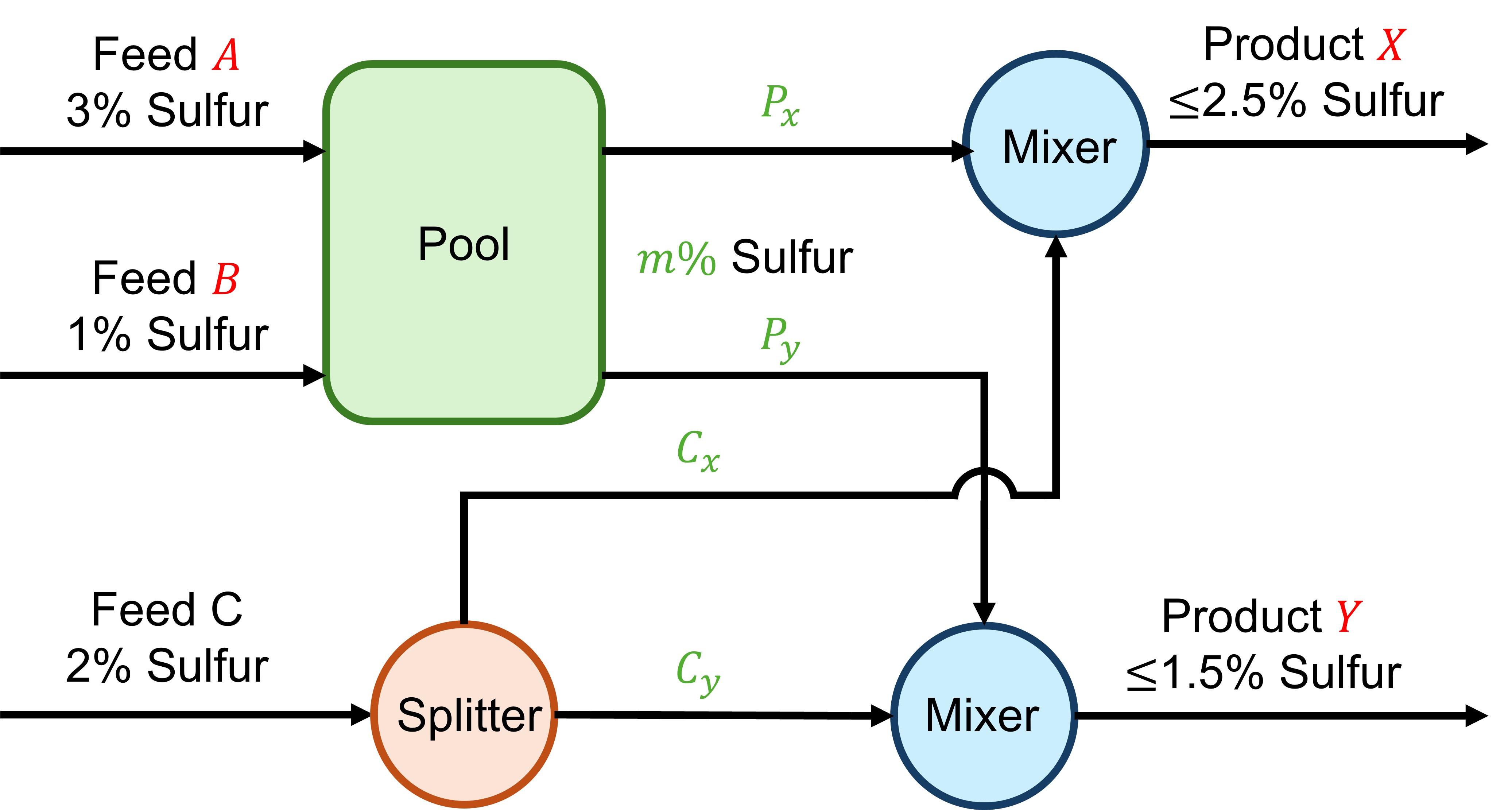}
   \caption{Process flowsheet of a pooling problem. The product can be formed from three feeds. The input variables are denoted in red. The output variables are denoted in green. Feed compositions are fixed. Feed A and B must be pooled together.}
   \label{fig:pooling_problem_flowsheet}
\end{figure}

We consider a pooling problem discussed in \cite{floudas2013handbook} where both equality and inequality constraints are present. The flowsheet is given in Figure \ref{fig:pooling_problem_flowsheet}, which consists of a pool, one splitter and two mixture. The product streams $X$ and $Y$ are produced by combining the feed steams $A$, $B$ and $C$ with sulfur content of 3\%, 1\% and 2\% respectively. The maximum sulfur content in product $X$ is 2.5\% and in product $Y$ is 1.5\% which yields the inequality constraints. The nonlinearity arises in the problem due to the fact that $A$ and $B$ must be pooled together. $m$ is the percentage of sulfur content of the streams coming out of the pool.

\subsubsection{Problem setup and data generation}
We vary four input variables: (1) feed flowrate $A$, (2) feed flowrate $B$, (3) product flowrate $X$, (4) product flowrate $Y$ and solved the system in python. The input space is sampled on a random grid: $A,B\in[0,500]$, $X\in[0,100]$ and $Y\in[0,200]$. The corresponding outputs include: (1) Percent amount of sulfur in the streams coming out of the pool ($m$), (2) flowrates of the streams coming from pool ($P_x, P_y$) and (3) splitted flowrates of Feed $C$ to the mixers ($C_x, C_y$). We generated 2000 data points for training the model.

\subsubsection{Nonlinear constraints: Equality and Inequality}
We consider four equality constraints derived from the material balances and two inequality constraints derived from the product specification requirements regarding the sulfur content.

\begin{enumerate}[label=(C\arabic*), leftmargin=2.2cm]
    \item Pool mass balance (affine equality):
    \begin{ceqn}
    \begin{equation}
    P_x + P_y = A + B
    \end{equation}
    \end{ceqn}

    \item Mixer mass balance for $X$ (affine equality): 
    \begin{ceqn}
    \begin{equation}
    X = P_x + C_x
    \end{equation}
    \end{ceqn}

    \item Mixer mass balance for $Y$(affine equality):
    \begin{ceqn}
    \begin{equation}
    Y = P_y + C_y
    \end{equation}
    \end{ceqn}

    \item Sulfur balance in pool (nonlinear equality):
    \begin{ceqn}
    \begin{equation}
    m P_x + m P_y = 3A + B
    \end{equation}
    \end{ceqn}

    \item Product Specification $X$ (nonlinear inequality):
    \begin{ceqn}
    \begin{equation}
    mP_x + 2C_x \leq 2.5X
    \end{equation}
    \end{ceqn}

    \item Product Specification $Y$ (nonlinear inequality):
    \begin{ceqn}
    \begin{equation}
    mP_y + 2C_y \leq 1.5X
    \end{equation}
    \end{ceqn}
\end{enumerate}

In this case study, we enforce the six algebraic constraints, both equality and inequality, by constructing the KKT system (without log-exponential transformation). In our implementation, we use $K = 100, \gamma = 10^{-2}$, and fixed $\alpha = 0.5$ (no backtracking).\\

\begin{figure}[t]
   \centering
   \includegraphics[width=1\linewidth]{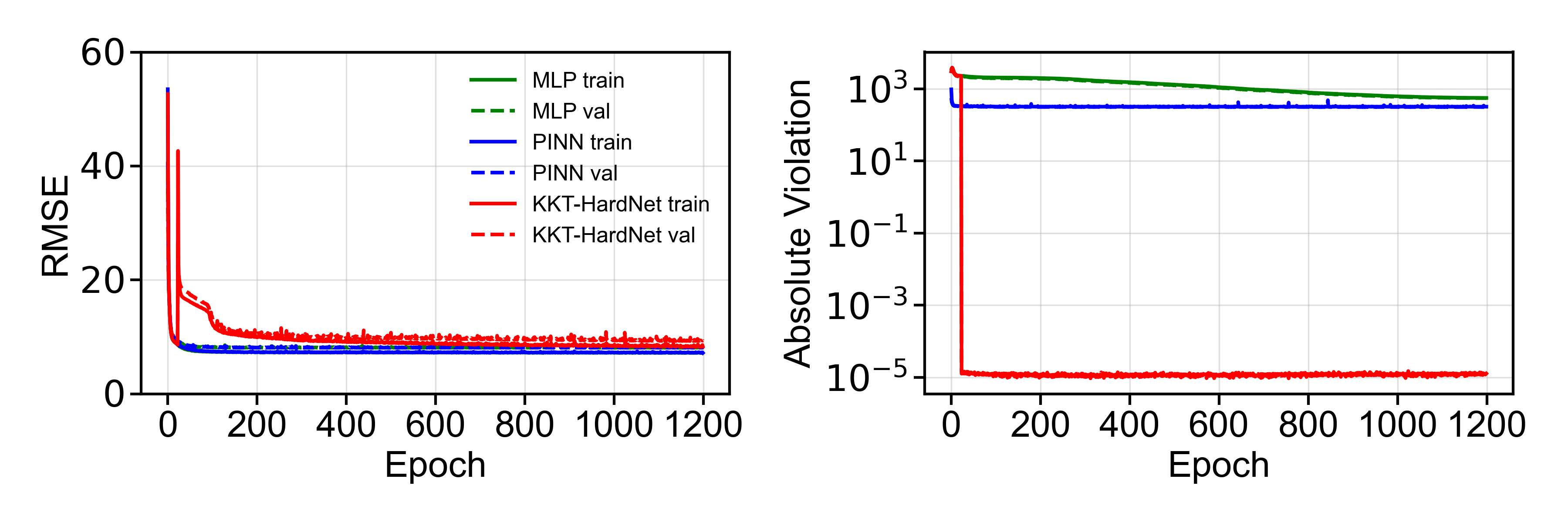}
   \caption{Learning curves for the pooling problem case study. Left: RMSE; Right: absolute constraint violation over 1200 epochs.}
   \label{fig:training_curves_pooling}
\end{figure}

We compare the performance of KKT-Hardnet with PINN and MLP, all using identical architectures and optimization settings (learning rate $10^{-3}$, 1200 epochs, Adam optimizer). The PINN model includes the constraints as a penalty term in the loss function with a penalty weight $\omega = 1.0$. The learning curves for RMSE and absolute constraint violation are provided in Figure \ref{fig:training_curves_pooling}. We observe constraint violations around $10^{-5}$. Note that the data was not standardized, which leads to a higher numerical precision error. Even in this case, compared to MLP and PINN, KKT-Hardnet reduces the constraint violation by 6-7 orders of magnitude.

Table \ref{tab:pooling_metrics} provides the MSE and absolute constraint violation for both training and testing data. Constraint satisfaction around $10^{-5}$ was achieved for KKT-Hardnet. The RMSE plot shows a spike around the 25th epoch for KKT-Hardnet, which is due to the activation of the projection layer, and eventually, the error drops close to MLP and PINN error. Though the MSE for KKT-Hardnet is higher than MLP and PINN, it is able to significantly reduce constraint violation.

\begin{table}[H]
\centering
\caption{Model performance comparison on the pooling problem case study.}
\label{tab:pooling_metrics}
\begin{tabular}{lcccc}
\toprule
& \multicolumn{2}{c}{\textbf{Training}} & \multicolumn{2}{c}{\textbf{Validation}}\\
\cmidrule(lr){2-3}\cmidrule(lr){4-5}
Model & MSE & \(|h| + |g|\) & MSE & \(|h| + |g|\) \\
\midrule
MLP & \(53.537\) & \(5.58\!\times10^{2}\) & \(67.758\) & \(5.51\!\times\!10^{2}\) \\
PINN & \(56.69\) & \(3.00\!\times\!10^{1}\) & \(71.32\) & \(2.94\!\times\!10^{1}\) \\
KKT-Hardnet & \(74.92\) & \(1.08\!\times\!10^{-5}\) & \(92.257\) & \(1.05\!\times\!10^{-5}\) \\
\bottomrule
\end{tabular}
\end{table}

\section{Conclusions}
\label{sec:conclusions}
We have presented a physics-informed neural network architecture that enforces hard nonlinear equality and inequality constraints in both inputs and outputs through a differentiable projection layer. The projection involves solving a square system of nonlinear equations corresponding to the KKT conditions of a distance minimization problem. The KKT system of equations is then subjected to a Newton-type iterative scheme, which acts as a differentiable projection layer within the neural network and guarantees constraint satisfaction. In doing so, we have addressed the limitations of traditional PINNs that rely on soft constraint penalty parameters and lack guaranteed satisfaction of first-principles-based governing equations. Furthermore, we have introduced a log-exponential transformation of a wide class of nonlinear equations (involving polynomial, power, rational, etc) to a structured form involving linear and exponential terms. Numerical experiments on illustrative examples, a pooling problem, and a highly nonlinear chemical process simulation problem demonstrate that the proposed approach eliminates constraint violations to within specified tolerance and machine precision. It also improves predictive accuracy compared to conventional MLPs and soft-constrained PINNs. Future directions include extending the method for application to differential-algebraic systems, learning the solutions of nonlinear optimization problem, and devising tailored algorithm that exploits the structure and sparsity of the Jacobian for the log-exponential system. To improve the initial guess of the Newton step at the projection layer, an unconstrained pre-training phase, followed by the activation of the projection layer, may be investigated.

\section*{Acknowledgments}
The authors gratefully acknowledge partial funding support from the NSF CAREER award CBET-1943479 and the EPA Project Grant 84097201. Part of the research was conducted with the computing resources provided by Texas A\&M High Performance Research Computing. 

\bibliographystyle{unsrt}  
\bibliography{references}  

\begin{thebibliography}{10}

\bibitem{karniadakis2021physics}
George~Em Karniadakis, Ioannis~G Kevrekidis, Lu~Lu, Paris Perdikaris, Sifan Wang, and Liu Yang.
\newblock Physics-informed machine learning.
\newblock {\em Nature Reviews Physics}, 3(6):422--440, 2021.

\bibitem{iftakher2025integrating}
Ashfaq Iftakher, Ty~Leonard, and M.~M~Faruque Hasan.
\newblock Integrating different fidelity models for process optimization: A case of equilibrium and rate-based extractive distillation using ionic liquids.
\newblock {\em Computers \& Chemical Engineering}, 192:108890, 2025.

\bibitem{eivazi2022physics}
Hamidreza Eivazi, Mojtaba Tahani, Philipp Schlatter, and Ricardo Vinuesa.
\newblock Physics-informed neural networks for solving reynolds-averaged navier--stokes equations.
\newblock {\em Physics of Fluids}, 34(7), 2022.

\bibitem{chen2021theory}
Yuntian Chen, Dou Huang, Dongxiao Zhang, Junsheng Zeng, Nanzhe Wang, Haoran Zhang, and Jinyue Yan.
\newblock Theory-guided hard constraint projection (hcp): A knowledge-based data-driven scientific machine learning method.
\newblock {\em Journal of Computational Physics}, 445:110624, 2021.

\bibitem{raissi2020hidden}
Maziar Raissi, Alireza Yazdani, and George~Em Karniadakis.
\newblock Hidden fluid mechanics: Learning velocity and pressure fields from flow visualizations.
\newblock {\em Science}, 367(6481):1026--1030, 2020.

\bibitem{sanchez2018real}
Carlos S{\'a}nchez-S{\'a}nchez and Dario Izzo.
\newblock Real-time optimal control via deep neural networks: study on landing problems.
\newblock {\em Journal of Guidance, Control, and Dynamics}, 41(5):1122--1135, 2018.

\bibitem{kumar2021industrial}
Pratyush Kumar, James~B Rawlings, and Stephen~J Wright.
\newblock Industrial, large-scale model predictive control with structured neural networks.
\newblock {\em Computers \& chemical engineering}, 150:107291, 2021.

\bibitem{BHOSEKAR2018}
Atharv Bhosekar and Marianthi Ierapetritou.
\newblock Advances in surrogate based modeling, feasibility analysis, and optimization: A review.
\newblock {\em Computers \& Chemical Engineering}, 108:250--267, 2018.

\bibitem{iftakher2022data}
Ashfaq Iftakher, Chinmay~M Aras, Mohammed~Sadaf Monjur, and M.~M.~Faruque Hasan.
\newblock Data-driven approximation of thermodynamic phase equilibria.
\newblock {\em AIChE Journal}, 68(6):e17624, 2022.

\bibitem{iftakher2022guaranteed}
Ashfaq Iftakher, Chinmay~M Aras, Mohammed~Sadaf Monjur, and M.~M.~Faruque Hasan.
\newblock Guaranteed error-bounded surrogate modeling and application to thermodynamics.
\newblock In {\em Computer Aided Chemical Engineering}, volume~49, pages 1831--1836. Elsevier, 2022.

\bibitem{Mcbride2019}
Kevin McBride and Kai Sundmacher.
\newblock Overview of surrogate modeling in chemical process engineering.
\newblock {\em Chemie Ingenieur Technik}, 91(3):228--239, 2019.

\bibitem{misener2023formulating}
Ruth Misener and Lorenz Biegler.
\newblock Formulating data-driven surrogate models for process optimization.
\newblock {\em Computers \& Chemical Engineering}, 179:108411, 2023.

\bibitem{wilson2017alamo}
Zachary~T Wilson and Nikolaos~V Sahinidis.
\newblock The alamo approach to machine learning.
\newblock {\em Computers \& Chemical Engineering}, 106:785--795, 2017.

\bibitem{cozad2014learning}
Alison Cozad, Nikolaos~V Sahinidis, and David~C Miller.
\newblock Learning surrogate models for simulation-based optimization.
\newblock {\em AIChE Journal}, 60(6):2211--2227, 2014.

\bibitem{hornik1989multilayer}
Kurt Hornik, Maxwell Stinchcombe, and Halbert White.
\newblock Multilayer feedforward networks are universal approximators.
\newblock {\em Neural networks}, 2(5):359--366, 1989.

\bibitem{krizhevsky2017imagenet}
Alex Krizhevsky, Ilya Sutskever, and Geoffrey~E Hinton.
\newblock Imagenet classification with deep convolutional neural networks.
\newblock {\em Communications of the ACM}, 60(6):84--90, 2017.

\bibitem{goodfellow2020generative}
Ian Goodfellow, Jean Pouget-Abadie, Mehdi Mirza, Bing Xu, David Warde-Farley, Sherjil Ozair, Aaron Courville, and Yoshua Bengio.
\newblock Generative adversarial networks.
\newblock {\em Communications of the ACM}, 63(11):139--144, 2020.

\bibitem{lecun2015deep}
Yann LeCun, Yoshua Bengio, and Geoffrey Hinton.
\newblock Deep learning.
\newblock {\em nature}, 521(7553):436--444, 2015.

\bibitem{vaswani2017attention}
Ashish Vaswani, Noam Shazeer, Niki Parmar, Jakob Uszkoreit, Llion Jones, Aidan~N Gomez, {\L}ukasz Kaiser, and Illia Polosukhin.
\newblock Attention is all you need.
\newblock {\em Advances in neural information processing systems}, 30, 2017.

\bibitem{ren2022tutorial}
Yi~Ming Ren, Mohammed~S Alhajeri, Junwei Luo, Scarlett Chen, Fahim Abdullah, Zhe Wu, and Panagiotis~D Christofides.
\newblock A tutorial review of neural network modeling approaches for model predictive control.
\newblock {\em Computers \& Chemical Engineering}, 165:107956, 2022.

\bibitem{cai2021physicsa}
Shengze Cai, Zhiping Mao, Zhicheng Wang, Minglang Yin, and George~Em Karniadakis.
\newblock Physics-informed neural networks (pinns) for fluid mechanics: A review.
\newblock {\em Acta Mechanica Sinica}, 37(12):1727--1738, 2021.

\bibitem{iftakher2025multi}
Ashfaq Iftakher, Mohammed~Sadaf Monjur, Ty~Leonard, Rafiqul Gani, and MM~Faruque Hasan.
\newblock Multiscale high-throughput screening of ionic liquid solvents for mixed-refrigerant separation.
\newblock {\em Computers \& Chemical Engineering}, 199:109138, 2025.

\bibitem{kashinath2021physics}
Karthik Kashinath, M~Mustafa, Adrian Albert, JL~Wu, C~Jiang, Soheil Esmaeilzadeh, Kamyar Azizzadenesheli, R~Wang, Ashesh Chattopadhyay, A~Singh, et~al.
\newblock Physics-informed machine learning: case studies for weather and climate modelling.
\newblock {\em Philosophical Transactions of the Royal Society A}, 379(2194):20200093, 2021.

\bibitem{raissi2019physics}
Maziar Raissi, Paris Perdikaris, and George~E Karniadakis.
\newblock Physics-informed neural networks: A deep learning framework for solving forward and inverse problems involving nonlinear partial differential equations.
\newblock {\em Journal of Computational physics}, 378:686--707, 2019.

\bibitem{cai2021physics}
Shengze Cai, Zhicheng Wang, Sifan Wang, Paris Perdikaris, and George~Em Karniadakis.
\newblock Physics-informed neural networks for heat transfer problems.
\newblock {\em Journal of Heat Transfer}, 143(6):060801, 2021.

\bibitem{chen2024physics}
Hao Chen, Gonzalo E~Constante Flores, and Can Li.
\newblock Physics-informed neural networks with hard linear equality constraints.
\newblock {\em Computers \& Chemical Engineering}, 189:108764, 2024.

\bibitem{wang2022and}
Sifan Wang, Xinling Yu, and Paris Perdikaris.
\newblock When and why pinns fail to train: A neural tangent kernel perspective.
\newblock {\em Journal of Computational Physics}, 449:110768, 2022.

\bibitem{ma2022data}
Kaiwen Ma, Nikolaos~V Sahinidis, Satyajith Amaran, Rahul Bindlish, Scott~J Bury, Devin Griffith, and Sreekanth Rajagopalan.
\newblock Data-driven strategies for optimization of integrated chemical plants.
\newblock {\em Computers \& Chemical Engineering}, 166:107961, 2022.

\bibitem{krishnapriyan2021characterizing}
Aditi Krishnapriyan, Amir Gholami, Shandian Zhe, Robert Kirby, and Michael~W Mahoney.
\newblock Characterizing possible failure modes in physics-informed neural networks.
\newblock {\em Advances in neural information processing systems}, 34:26548--26560, 2021.

\bibitem{wang20222}
Chuwei Wang, Shanda Li, Di~He, and Liwei Wang.
\newblock Is \ensuremath{L^2} physics-informed loss always suitable for training physics-informed neural networks?
\newblock {\em Advances in Neural Information Processing Systems}, 35:8278--8290, 2022.

\bibitem{rathore2024challenges}
Pratik Rathore, Weimu Lei, Zachary Frangella, Lu~Lu, and Madeleine Udell.
\newblock Challenges in training pinns: A loss landscape perspective.
\newblock {\em arXiv preprint arXiv:2402.01868}, 2024.

\bibitem{min2024hard}
Youngjae Min, Anoopkumar Sonar, and Navid Azizan.
\newblock Hard-constrained neural networks with universal approximation guarantees.
\newblock {\em arXiv preprint arXiv:2410.10807}, 2024.

\bibitem{marquez2017imposing}
Pablo M{\'a}rquez-Neila, Mathieu Salzmann, and Pascal Fua.
\newblock Imposing hard constraints on deep networks: Promises and limitations.
\newblock {\em arXiv preprint arXiv:1706.02025}, 2017.

\bibitem{beucler2021enforcing}
Tom Beucler, Michael Pritchard, Stephan Rasp, Jordan Ott, Pierre Baldi, and Pierre Gentine.
\newblock Enforcing analytic constraints in neural networks emulating physical systems.
\newblock {\em Physical review letters}, 126(9):098302, 2021.

\bibitem{donti2021dc3}
Priya~L Donti, David Rolnick, and J~Zico Kolter.
\newblock Dc3: A learning method for optimization with hard constraints.
\newblock {\em arXiv preprint arXiv:2104.12225}, 2021.

\bibitem{nocedal1999numerical}
Jorge Nocedal and Stephen~J Wright.
\newblock {\em Numerical optimization}.
\newblock Springer, 1999.

\bibitem{amos2017optnet}
Brandon Amos and J~Zico Kolter.
\newblock Optnet: Differentiable optimization as a layer in neural networks.
\newblock In {\em International conference on machine learning}, pages 136--145. PMLR, 2017.

\bibitem{agrawal2019differentiable}
Akshay Agrawal, Brandon Amos, Shane Barratt, Stephen Boyd, Steven Diamond, and J~Zico Kolter.
\newblock Differentiable convex optimization layers.
\newblock {\em Advances in neural information processing systems}, 32, 2019.

\bibitem{lastrucci2025enforce}
Giacomo Lastrucci and Artur~M Schweidtmann.
\newblock Enforce: Exact nonlinear constrained learning with adaptive-depth neural projection.
\newblock {\em arXiv preprint arXiv:2502.06774}, 2025.

\bibitem{jiang1997smoothed}
Houyuan Jiang.
\newblock Smoothed fischer-burmeister equation methods for the complementarity problem.
\newblock {\em Department of Mathematics, The University of Melbourne (Australia}, 1997.

\bibitem{paszke2019pytorch}
A~Paszke.
\newblock Pytorch: An imperative style, high-performance deep learning library.
\newblock {\em arXiv preprint arXiv:1912.01703}, 2019.

\bibitem{kingma2014adam}
Diederik~P Kingma.
\newblock Adam: A method for stochastic optimization.
\newblock {\em arXiv preprint arXiv:1412.6980}, 2014.

\bibitem{barata2012moore}
Jo{\~a}o Carlos~Alves Barata and Mahir~Saleh Hussein.
\newblock The moore--penrose pseudoinverse: A tutorial review of the theory.
\newblock {\em Brazilian Journal of Physics}, 42:146--165, 2012.

\bibitem{monjur2022separation}
Mohammed~Sadaf Monjur, Ashfaq Iftakher, and M.~M.~Faruque Hasan.
\newblock Separation process synthesis for high-gwp refrigerant mixtures: Extractive distillation using ionic liquids.
\newblock {\em Industrial \& Engineering Chemistry Research}, 61(12):4390--4406, 2022.

\bibitem{floudas2013handbook}
Christodoulos~A Floudas, Panos~M Pardalos, Claire Adjiman, William~R Esposito, Zeynep~H G{\"u}m{\"u}s, Stephen~T Harding, John~L Klepeis, Clifford~A Meyer, and Carl~A Schweiger.
\newblock {\em Handbook of test problems in local and global optimization}, volume~33.
\newblock Springer Science \& Business Media, 2013.

\end{thebibliography}

\newpage

\section*{Appendix~A: Effect of the penalty weight on PINN training}

\label{sec:pinn-lambda}

We study the influence of the soft-constraint penalty weight $w$ in PINN for the extractive distillation case study described in Section \ref{sec:ED-column}. We rerun the experiments and report a shaded band across $w\in [0.1, 100]$ sweep showing the epoch-wise min--max envelope with the mean curve for both RMSE and mean absolute violation \(|h|\). Note that the exact MSE and violation values may slightly differ across different random seeds because the training does not follow the same optimization path due to the nature of the stochastic gradient descent. Also, the DataLoader shuffling in different runs can yield a different minibatch order, which may prompt different weights in each epoch, resulting in a slightly different Newton‐projection convergence. However, as described next, the conclusions are consistent.

\begin{figure}[t]
   \centering
   \includegraphics[width=1\linewidth]{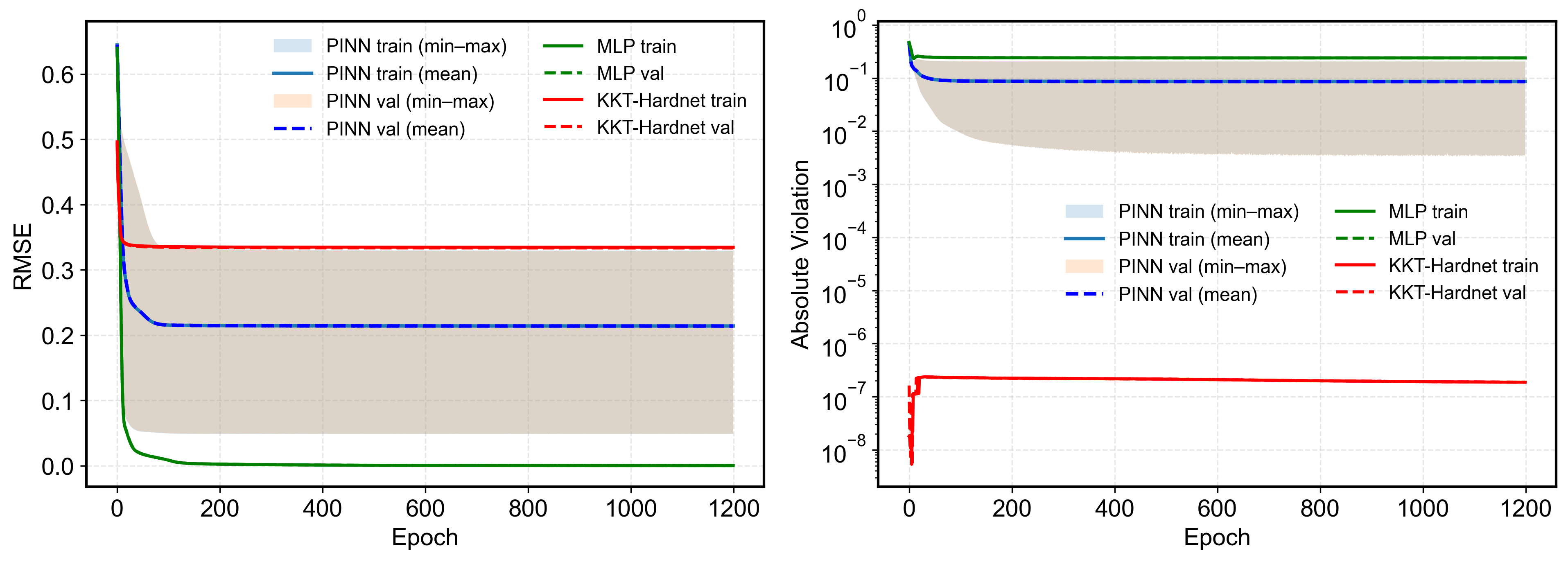}
   \caption{Training curves comparing MLP, KKT-Hardnet, and PINNs with a sweep over $w \in [0.1, 100]$. Left: RMSE; right: mean absolute constraint violation. Shaded regions show the min--max envelope across $w$ with a solid (train) and dashed (val) mean.}
   \label{fig:pinn-lambda-case1}
\end{figure}

\paragraph{Nonlinear constraints in both input and output.}
This system corresponds to the problem definition and constraints described in Section \ref{Sec:nonlinear-ED}. The unconstrained MLP achieves low data error but violates known constraints substantially (val MSE $\approx 4.3\times10^{-8}$ versus \(|h|\approx 2.39\times10^{-1}\)). KKT-Hardnet attains near-zero violation (\(|h|\approx 1.9\times10^{-7}\)) with moderate data error (val MSE $\approx 1.11\times10^{-1}$). The PINN (as shown in Fig \ref{fig:pinn-lambda-case1}) exhibits the expected Pareto trade-off. As $w$ increases, violation decreases monotonically while MSE increases. With $w=0.1$ we obtain the best validation MSE ($2.40\times10^{-3}$) but poor constraint satisfaction (\(|h|=2.05\times10^{-1}\)). On the other hand, with $w=100$ the violation drops to \(|h|=3.42\times10^{-3}\) at the cost of an increased validation MSE $1.08\times10^{-1}$. Detailed results are given in \emph{Table~\ref{tab:case1_lambda}}.

\begin{table}[H]
\centering
\caption{PINN performance across different $w$ for the extractive distillation case study with nonlinear input-output constraints.}
\label{tab:case1_lambda}
\begin{tabular}{lcccc}
\toprule
& \multicolumn{2}{c}{\textbf{Training}} & \multicolumn{2}{c}{\textbf{Validation}}\\
\cmidrule(lr){2-3}\cmidrule(lr){4-5}
Model & MSE & \(|h|\) & MSE & \(|h|\) \\
\midrule
MLP & \(3.390\times10^{-8}\) & \(2.39\times10^{-1}\) &
      \(4.302\times10^{-8}\) & \(2.39\times10^{-1}\) \\
KKT-Hardnet & \(1.120\times10^{-1}\) & \(1.87\times10^{-7}\) &
              \(1.112\times10^{-1}\) & \(1.87\times10^{-7}\) \\
\midrule
PINN $\lambda{=}0.1$   & \(2.368\times10^{-3}\) & \(2.05\times10^{-1}\) &
                         \(2.397\times10^{-3}\) & \(2.05\times10^{-1}\) \\
PINN $\lambda{=}1$     & \(3.355\times10^{-2}\) & \(1.10\times10^{-1}\) &
                         \(3.377\times10^{-2}\) & \(1.09\times10^{-1}\) \\
PINN $\lambda{=}10$    & \(8.674\times10^{-2}\) & \(2.57\times10^{-2}\) &
                         \(8.656\times10^{-2}\) & \(2.54\times10^{-2}\) \\
PINN $\lambda{=}100$   & \(1.085\times10^{-1}\) & \(3.42\times10^{-3}\) &
                         \(1.079\times10^{-1}\) & \(3.42\times10^{-3}\) \\
\bottomrule
\end{tabular}
\end{table}

\begin{figure}[t]
   \centering
   \includegraphics[width=1\linewidth]{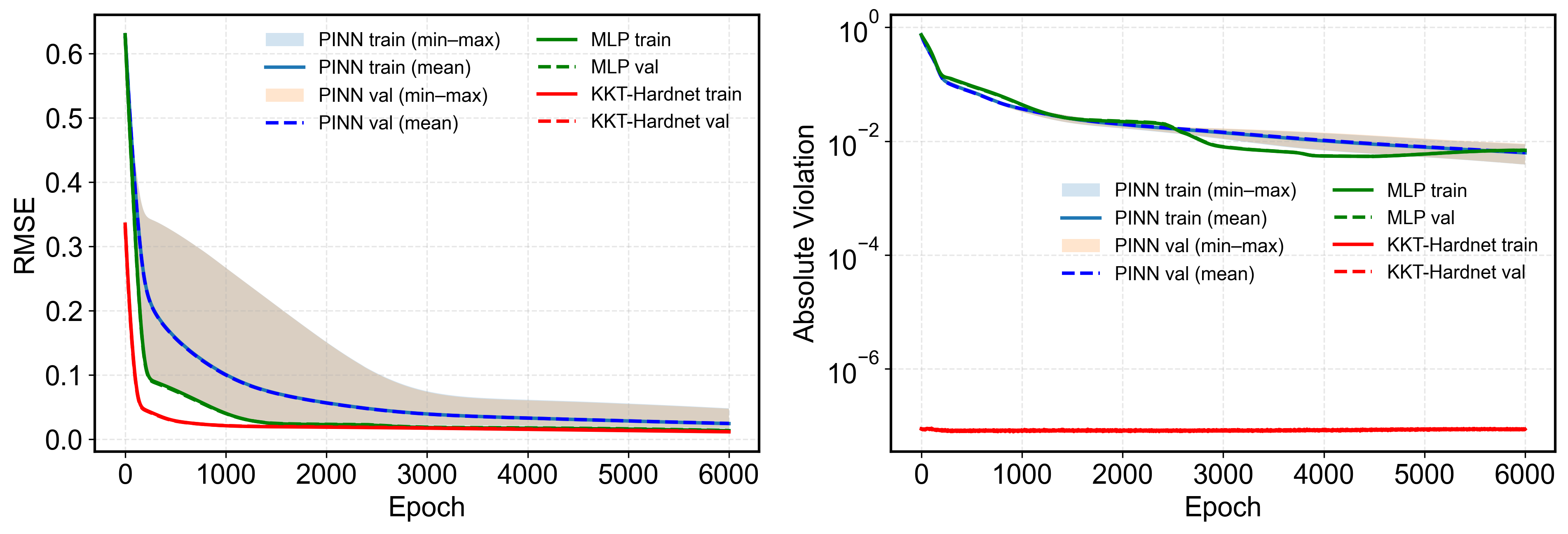}
   \caption{Training curves comparing MLP, KKT-Hardnet, and PINNs with a sweep over $w \in [0.1, 100]$. Left: RMSE; right: mean absolute constraint violation. Shaded regions show the min--max envelope across $w$ with a solid (train) and dashed (val) mean.}
   \label{fig:pinn-lambda-case2}
\end{figure}

\paragraph{Affine-in-output constraints.}
This system corresponds to the problem definition and constraints described in Section \ref{sec:linear-ED}. Here, the effect of $w$ is not pronounced, and the spread across $w$ is small in the final epochs (see the narrow violation band in \emph{Fig.~\ref{fig:pinn-lambda-case2}}). KKT-Hardnet again achieves negligible violation (\(\sim 8.6\times10^{-8}\)) and, notably, better data fit than the MLP (validation MSE $1.48\times10^{-4}$ vs. $1.93\times10^{-4}$). For the PINN, the best validation MSE occurs at $w=0.1$ while the lowest violation is reached near $w=1$ (see \emph{Table~\ref{tab:case2_lambda}}). This case thus highlights the practical difficulty of choosing a single penalty weight for PINN that simultaneously optimizes data fit and constraint feasibility. Soft penalties in PINN require tuning $w$ to navigate a Pareto front that is problem-dependent.

\begin{table}[H]
\centering
\caption{PINN performance across different $w$ for the extractive distillation case study with affine-in-output constraints.}
\label{tab:case2_lambda}
\begin{tabular}{lcccc}
\toprule
& \multicolumn{2}{c}{\textbf{Training}} & \multicolumn{2}{c}{\textbf{Validation}}\\
\cmidrule(lr){2-3}\cmidrule(lr){4-5}
Model & MSE & \(|h|\) & MSE & \(|h|\) \\
\midrule
MLP & \(1.828\times10^{-4}\) & \(6.89\times10^{-3}\) &
      \(1.932\times10^{-4}\) & \(6.90\times10^{-3}\) \\
KKT-Hardnet & \(1.390\times10^{-4}\) & \(8.66\times10^{-8}\) &
              \(1.479\times10^{-4}\) & \(8.61\times10^{-8}\) \\
\midrule
PINN $\lambda{=}0.1$  & \(2.899\times10^{-4}\) & \(8.79\times10^{-3}\) &
                        \(3.061\times10^{-4}\) & \(8.96\times10^{-3}\) \\
PINN $\lambda{=}0.3$  & \(2.906\times10^{-4}\) & \(6.72\times10^{-3}\) &
                        \(3.066\times10^{-4}\) & \(6.84\times10^{-3}\) \\
\textbf{PINN} $\boldsymbol{\lambda{=}1}$    & \(3.056\times10^{-4}\) & \(\boldsymbol{3.85}\times \boldsymbol{10^{-3}}\) &
                        \(3.221\times10^{-4}\) & \(\boldsymbol{3.88}\times \boldsymbol{10^{-3}}\) \\
PINN $\lambda{=}3$    & \(4.163\times10^{-4}\) & \(5.92\times10^{-3}\) &
                        \(4.347\times10^{-4}\) & \(6.02\times10^{-3}\) \\
PINN $\lambda{=}10$   & \(4.886\times10^{-4}\) & \(5.80\times10^{-3}\) &
                        \(5.062\times10^{-4}\) & \(5.89\times10^{-3}\) \\
PINN $\lambda{=}30$   & \(8.743\times10^{-4}\) & \(6.17\times10^{-3}\) &
                        \(8.756\times10^{-4}\) & \(6.22\times10^{-3}\) \\
PINN $\lambda{=}100$  & \(2.340\times10^{-3}\) & \(6.28\times10^{-3}\) &
                        \(2.271\times10^{-3}\) & \(6.36\times10^{-3}\) \\
\bottomrule
\end{tabular}
\end{table}

\section*{Appendix~B: Implicit differentiation for KKT-Hardnet}

Instead of backpropagating through all Newton steps, one can differentiate the solution of the KKT system directly. To illustrate this, recall that the KKT residuals are collected as

\begin{equation}
\label{eq:F-def}
\boldsymbol F(\boldsymbol y,\boldsymbol\lambda;\boldsymbol x,\hat{\boldsymbol y}) =
\begin{bmatrix}
\boldsymbol y-\hat{\boldsymbol y}
+ \nabla_{\boldsymbol y}\boldsymbol{\tilde h}(\boldsymbol x,\boldsymbol y,\boldsymbol\lambda)^{\!\top}\boldsymbol\lambda\\[4pt]
\boldsymbol{\tilde h}(\boldsymbol x,\boldsymbol y,\boldsymbol\lambda)
\end{bmatrix}
=\boldsymbol 0,
\end{equation}

where \(\boldsymbol{\tilde h}\) contains both equalities and the inequality reformulation. Define

\[
\boldsymbol J_s := \frac{\partial \boldsymbol F}{\partial(\boldsymbol y,\boldsymbol\lambda)},
\qquad
\boldsymbol J_{\hat y} := \frac{\partial \boldsymbol F}{\partial \hat{\boldsymbol y}},
\qquad
\boldsymbol J_x := \frac{\partial \boldsymbol F}{\partial \boldsymbol x}.
\]

Differentiating \(\boldsymbol F=0\) with respect to any variable \(t\in\{\boldsymbol\Theta,\boldsymbol x\}\) gives

\begin{equation}
\label{eq:implicit-diff}
\boldsymbol J_s
\begin{bmatrix}
\dfrac{\partial \tilde{\boldsymbol y}}{\partial t}\\[4pt]
\dfrac{\partial \tilde{\boldsymbol\lambda}}{\partial t}
\end{bmatrix}
\;+\;
\boldsymbol J_{\hat y}\,\frac{\partial \hat{\boldsymbol y}}{\partial t}
\;+\;
\boldsymbol J_x\,\frac{\partial \boldsymbol x}{\partial t}
\;=\;\boldsymbol 0.
\end{equation}

By the chain rule,
\(
\dfrac{\mathrm d \mathcal L}{\mathrm d t}
=
\bigl(\tfrac{\partial \mathcal L}{\partial \tilde{\boldsymbol y}}\bigr)^{\!\top}
\dfrac{\partial \tilde{\boldsymbol y}}{\partial t}.
\)

Introduce the adjoint \(\boldsymbol v=[\boldsymbol v_y;\boldsymbol v_\lambda]\) as the solution of the transpose–KKT system

\begin{equation}
\label{eq:adjoint}
\boldsymbol J_s^{\!\top}\,\boldsymbol v
=
\begin{bmatrix}
\dfrac{\partial \mathcal L}{\partial \tilde{\boldsymbol y}}\\[2pt]
\boldsymbol 0
\end{bmatrix}.
\end{equation}

Left-multiplying \eqref{eq:implicit-diff} by \(\boldsymbol v^{\!\top}\) yields the vector–Jacobian product (VJP) form

\begin{equation}
\label{eq:total-deriv}
\frac{\mathrm d \mathcal L}{\mathrm d t}
=
-\,\boldsymbol v^{\!\top}
\Bigl(
\boldsymbol J_{\hat y}\,\frac{\partial \hat{\boldsymbol y}}{\partial t}
+
\boldsymbol J_x\,\frac{\partial \boldsymbol x}{\partial t}
\Bigr).
\end{equation}

In our formulation, the stationarity block \eqref{eq:F-def} contains \(+\boldsymbol y-\hat{\boldsymbol y}\), hence
\[
\boldsymbol J_{\hat y}
=
\frac{\partial}{\partial \hat{\boldsymbol y}}
\begin{bmatrix}
\boldsymbol y-\hat{\boldsymbol y}\\[2pt]
\boldsymbol 0
\end{bmatrix}
=
\begin{bmatrix}
-\boldsymbol I\\[2pt]\boldsymbol 0
\end{bmatrix}.
\]

Therefore,
\begin{equation}\label{eq:param-grad}
\frac{\mathrm d\mathcal L}{\mathrm d\boldsymbol\Theta}
=
\boldsymbol v_y^{\!\top}\,\frac{\partial \hat{\boldsymbol y}}{\partial\boldsymbol\Theta}
\qquad\text{and}\qquad
\frac{\mathrm d\mathcal L}{\mathrm d\boldsymbol x}
=
-\,\boldsymbol v^{\!\top}\boldsymbol J_x
\;+\;
\boldsymbol v_y^{\!\top}\,\frac{\partial \hat{\boldsymbol y}}{\partial \boldsymbol x}.
\end{equation}

\paragraph{Numerical and implementation notes.} \emph{Unrolled Newton/Gauss–Newton} lets autograd backpropagate through every iteration and linear solve. This is simple but memory-heavy, because all intermediate states of the $K$ iterations must be kept for the backward pass. The \emph{Implicit adjoint} method discussed above is \textbf{memory-light}. This is because, at the backward pass, only the single transpose–KKT system is solved, which is then backpropagated through the MLP backbone. If the forward projection uses a ridge \(\gamma>0\) in the normal equations, using the same \(\gamma\) in the adjoint (e.g., solving \((\boldsymbol J_s^{\!\top}\boldsymbol J_s+\gamma\boldsymbol I)\,\boldsymbol v_y=\partial \mathcal L/\partial \tilde{\boldsymbol y}\) after eliminating \(\boldsymbol v_\lambda\)) may improve robustness for nearly singular constraints.

\end{document}